\documentclass{article}

\usepackage{amsmath,amsfonts,bm}

\newcommand{\dataset}[1]{\mathcal{D}^\text{#1}}
\newcommand{\st}[0]{\text{s.t.}}

\def\eqref#1{equation~\ref{#1}}

\def\1{\bm{1}}

\def\vlambda{\bm{\lambda}}

\def\vtheta{{\bm{\theta}}}

\def\va{{\bm{a}}}
\def\vb{{\bm{b}}}

\def\ve{{\bm{e}}}
\def\vf{{\bm{f}}}
\def\vg{{\bm{g}}}

\def\vp{{\bm{p}}}

\def\vs{{\bm{s}}}

\def\vv{{\bm{v}}}
\def\vw{{\bm{w}}}
\def\vx{{\bm{x}}}
\def\vy{{\bm{y}}}
\def\vz{{\bm{z}}}

\def\mA{{\bm{A}}}
\def\mB{{\bm{B}}}

\def\mF{{\bm{F}}}
\def\mG{{\bm{G}}}
\def\mH{{\bm{H}}}
\def\mI{{\bm{I}}}
\def\mJ{{\bm{J}}}

\def\mR{{\bm{R}}}
\def\mS{{\bm{S}}}

\def\mV{{\bm{V}}}
\def\mW{{\bm{W}}}
\def\mX{{\bm{X}}}

\DeclareMathAlphabet{\mathsfit}{\encodingdefault}{\sfdefault}{m}{sl}
\SetMathAlphabet{\mathsfit}{bold}{\encodingdefault}{\sfdefault}{bx}{n}

\def\gD{{\mathcal{D}}}

\def\gJ{{\mathcal{J}}}

\def\gL{{\mathcal{L}}}

\def\sR{{\mathbb{R}}}

\newcommand{\E}{\mathbb{E}}

\newcommand{\softmax}{\mathrm{softmax}}

\DeclareMathOperator*{\argmin}{arg\,min}

\usepackage{nicefrac} 
\newcommand{\flatten}{\operatorname{vec}}

\usepackage[textsize=scriptsize]{todonotes}

\usepackage{bbding}
\usepackage{xspace}
\newcommand*{\ie}{i.e.\@\xspace}
\newcommand*{\iid}{i.i.d.\@\xspace}
\newcommand*{\wrt}{w.r.t.\@\xspace}
\newcommand*{\eg}{e.g.\@\xspace}

\newcommand{\outerf}{\mathcal{J}_{\text{out}}}

\ifx\BlackBox\undefined
\newcommand{\BlackBox}{\rule{1.5ex}{1.5ex}}  
\fi

\ifx\QED\undefined
\def\QED{~\rule[-1pt]{5pt}{5pt}\par\medskip}
\fi

\ifx\proof\undefined
\newenvironment{proof}{\par\noindent{\em Proof:\ }}{\hfill\BlackBox\\[.0mm]}
\fi

\ifx\theorem\undefined
\newtheorem{theorem}{Theorem}
\fi

\ifx\example\undefined
\newtheorem{example}{Example}
\fi

\ifx\property\undefined
\newtheorem{property}{Property}
\fi

\ifx\lemma\undefined
\newtheorem{lemma}{Lemma}
\fi

\ifx\proposition\undefined
\newtheorem{proposition}{Proposition}
\fi

\ifx\remark\undefined
\newtheorem{remark}{Remark}
\fi

\ifx\corollary\undefined
\newtheorem{corollary}{Corollary}
\fi

\ifx\definition\undefined
\newtheorem{definition}{Definition}
\fi

\ifx\conjecture\undefined
\newtheorem{conjecture}{Conjecture}
\fi

\ifx\axiom\undefined
\newtheorem{axiom}[theorem]{Axiom}
\fi

\ifx\claim\undefined
\newtheorem{claim}[theorem]{Claim}
\fi

\ifx\assumption\undefined
\newtheorem{assumption}{Assumption}
\fi

\newcommand{\papertitle}{Efficient Bilevel Optimization with KFAC-Based Hypergradients}

\usepackage[accepted]{aistats2026}
\usepackage{times}
\usepackage{hyperref}
\usepackage{cleveref}
\usepackage{xurl}
\usepackage{booktabs}
\usepackage{multirow}
\usepackage{multicol}
\usepackage{graphicx}
\usepackage{mathtools}
\usepackage{booktabs}
\usepackage{siunitx}
\usepackage{natbib}
\usepackage{multibib}
\usepackage{subcaption}
\usepackage{wrapfig}
\usepackage{algorithm}
\usepackage{algorithmic}
\usepackage{url}
\usepackage{cleveref}
\usepackage{enumitem}
\newcites{Appx}{Additional References}

\begin{document}

\twocolumn[

\aistatstitle{\papertitle}

\aistatsauthor{Disen Liao \And Felix Dangel \And Yaoliang Yu}

\aistatsaddress{ University of Waterloo \\ Vector Institute \And  Vector Institute \And University of Waterloo \\ Vector Institute } ]


\begin{abstract}
Bilevel optimization (BO) is widely applicable to many machine learning problems.
Scaling BO, however, requires repeatedly computing hypergradients, which involves solving inverse Hessian-vector products (IHVPs). In practice, these operations are often approximated using crude surrogates such as one-step gradient unrolling or identity/short Neumann expansions, which discard curvature information. 
We build on implicit function theorem-based algorithms and propose to incorporate Kronecker-factored approximate curvature (KFAC), yielding curvature-aware hypergradients with a better performance efficiency trade-off than Conjugate Gradient (CG) or Neumann methods and consistently outperforming unrolling. We evaluate this approach across diverse tasks, including meta-learning and AI safety problems.
On models up to BERT, we show that curvature information is valuable at scale, and KFAC can provide it with only modest memory and runtime overhead. Our implementation is available at \url{https://github.com/liaodisen/NeuralBo}.


\end{abstract}

\section{Introduction}
Bilevel optimization (BO), originally studied in economics \citep{von2010market}, models interactions between two decision-makers: a leader who acts first and a follower who optimizes in response. This leader-follower structure captures scenarios where one optimization problem depends on another. In machine learning, BO provides a unifying framework for diverse tasks: in meta-learning to acquire transferable inductive biases \citep{bertinetto2018meta, rajeswaran2019meta}, in hyper-parameter and data optimization to adapt learning pipelines \citep{franceschi2018bilevel, shu2019meta, feng2023embarassingly, hu2023meta, lorraine2020optimizing}, and in neural architecture search to guide model design \citep{liu2018darts}. BO
also plays a central role in AI safety, covering adversarial
problems such as data poisoning \citep{lu2022indiscriminate, radiya2021data, huang2020metapoison} and unlearnable examples \citep{huang2021unlearnable, liu2024game}. BO’s appeal
lies in its breadth: a single framework to reason about dependencies across learning, optimization, and safety.

Broadly speaking, algorithms for BO fall into three families: \emph{gradient unrolling} (GU), \emph{implicit function theorem} (IFT), and \emph{value-function} (VF). GU methods directly differentiate through a sequence of lower-level optimization steps \citep{franceschi2017forward, shaban2019truncated, shen2024memory}. While conceptually simple, they demand substantial memory and computation, and short unrolling horizons introduce truncation bias. IFT methods avoid explicit unrolling by leveraging the implicit function theorem to characterize the sensitivity of the lower-level solution \citep{lorraine2020optimizing, grazzi2020iteration, ji_bilevel_2021, choe2023making}. These approaches circumvent the need to store the intermediate iterates and are therefore more memory-efficient than unrolling, but hinge on computing inverse Hessian-vector products (IHVPs), which becomes a major computational bottleneck in high dimensions. VF methods instead sidestep second-order computations by reformulating BO as a constrained single-level problem \citep{liu2021value,  kwon2023fully, giovannelli2021bilevel}. Despite avoiding explicit second-order operations, their practical scalability remains less clear: solving the resulting single-level formulation often introduces additional optimization variables and constraints, which can increase complexity and make them less competitive in large-scale stochastic learning settings \citep{zhang2024introduction}.

Among the three, IFT-based approaches have recently gained popularity due to their ability to capture accurate hypergradients \citep{ghadimi2018approximation, ji_bilevel_2021}. Several efforts have pushed them toward larger-scale applications \citep{lorraine2020optimizing, choe2022betty, choe2023making}. Since exact inversion of the Hessian is infeasible for modern models, practitioners often resort to crude approximations. A common strategy truncates the Neumann series after a few terms \citep{lorraine2020optimizing, choe2022betty}. However, full convergence is usually computationally out of reach at this problem size, and the neural net's inner problem is frequently ill-conditioned \citep{sagun2016eigenvalues}, making short truncations inaccurate. Others go further by replacing the Hessian with the identity matrix \citep{choe2023making, hong_two-timescale_2022, liu2018darts}, effectively collapsing IFT into a one-step unrolling method and discarding curvature information altogether. As a result, existing methods either degrade under poor conditioning or rely on overly simplistic surrogates
that produce inaccurate hypergradients.

Our work builds
squarely within the IFT territory, but seeks to address its central bottleneck: scalable IHVPs. To this end, we explore Kronecker-Factored Approximate Curvature \citep[KFAC,][]{martens2015optimizing}, which provides
an efficient block-diagonal curvature approximation using the (uncentered) covariance of layer inputs and gradients.
Although rarely used in BO, recent large-scale results on influence functions in language models \citep{grosse2023studying} suggests that KFAC offers a practical middle ground between crude approximations and iterative curvature-aware methods. In contrast to influence functions, each outer step of IFT-based BO needs an IHVP, each of which is often obtained by iterative solvers via many HVPs. KFAC replaces these repeated HVPs with a structured curvature estimate, enabling efficient approximate inversion. 

\begin{figure}[!t]
    \centering
    \begin{subfigure}[b]{0.48\textwidth}
        \centering
        \includegraphics[width=\textwidth]{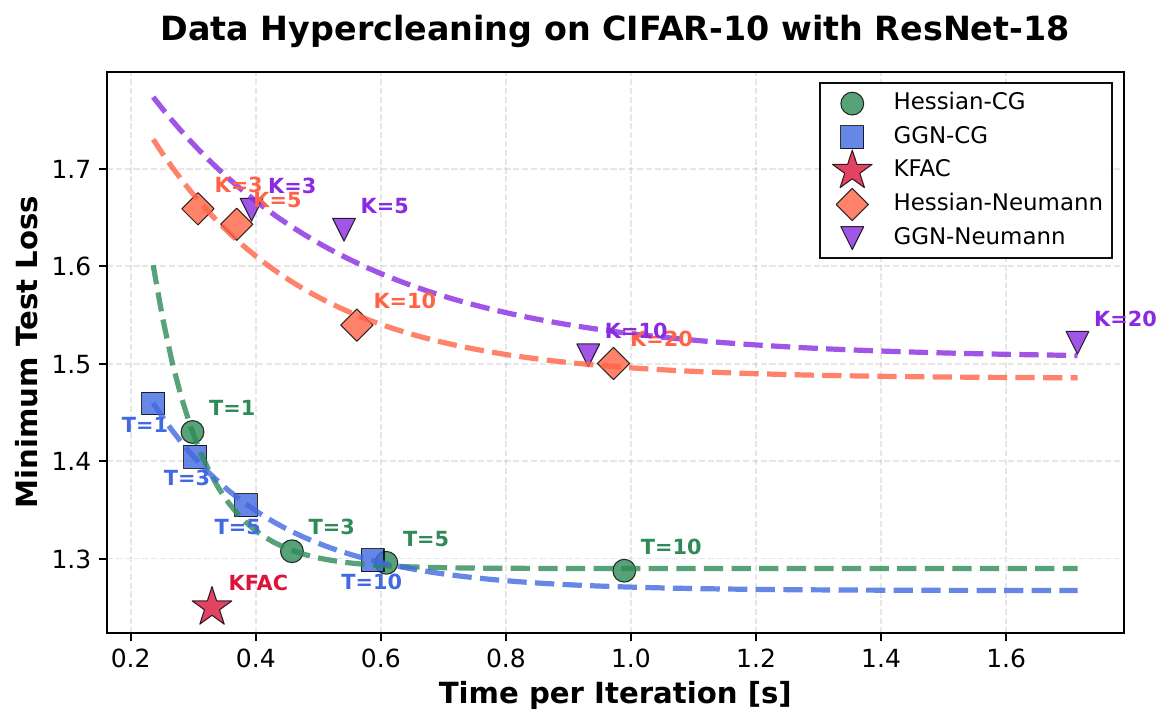}
        \label{fig:overparam_regime}
    \end{subfigure}
    \hfill 
    \vspace{-0.8cm}
    \begin{subfigure}{0.48\textwidth}
        \centering
        \includegraphics[width=\textwidth]{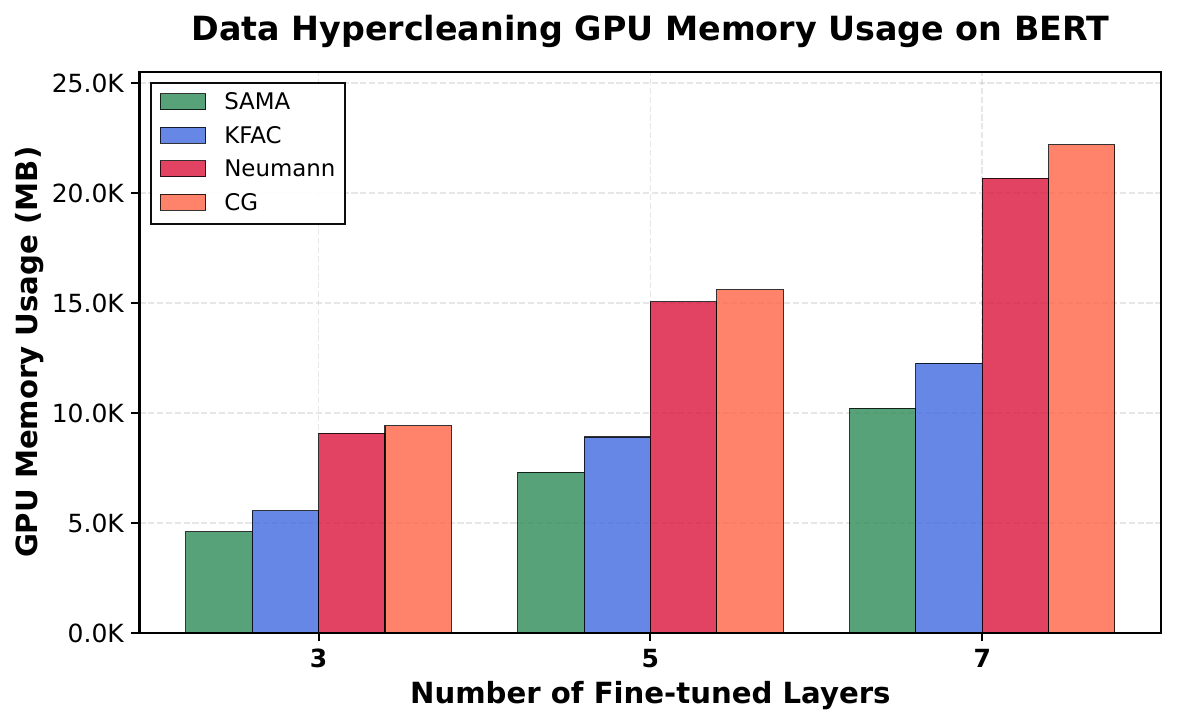}
        \label{fig:convergence_plot}
    \end{subfigure}
    \vspace{-0.9cm}
    \caption{\textbf{Top}: KFAC balances accuracy and efficiency
better than Hessian/GGN-based IHVP approximations. $T$ is the number of CG iterations and $K$ the number of truncated Neumann terms. \textbf{Bottom}: On BERT, KFAC incurs only a small memory overhead compared to the curvature-free  SAMA, while using less memory than CG/Neumann.}
    \label{fig:intro}
\end{figure}

We propose a KFAC-based hypergradient method for BO that directly plugs into existing algorithms. It provides rich curvature information at low computational cost---illustrated in \Cref{fig:intro} where KFAC reaches lower test loss with less per-iteration time than iterative IHVP approximations, with only modest memory overhead over HVP-based methods---and is effective on large-scale models across diverse applications. Our main contributions are:


\begin{itemize}
    \item We propose using inverse KFAC-vector products (IKVPs) for hypergradient computation in IFT-based BO and use them to replace IHVPs in existing methods, yielding a scalable curvature-aware approximation that improves wall-clock efficiency, convergence, and stability under small batch sizes.


    \item We scale data hypercleaning to BERT and show that incorporating curvature improves large-scale performance with only modest time and memory overhead.

    \item We demonstrate broad applicability by applying KFAC (including an empirical variant) to two meta-learning tasks and two ML safety problems, consistently improving performance across domains.
\end{itemize}

\section{Background}

\paragraph{Problem Setup.}
We formulate BO as a nested problem with an outer variable $\vlambda \in \mathbb{R}^m$ (e.g., hyperparameters) and an inner variable $\vtheta \in \mathbb{R}^d$ (e.g., model parameters). The goal is to minimize an outer objective $\mathcal{J}_{\text{out}}$ that depends on the optimal solution of the inner problem $\mathcal{J}_{\text{in}}$:
\begin{align*}
\min_{\vlambda \in \mathbb{R}^m} \; \Phi(\vlambda)
&:= \mathcal{J}_{\text{out}}\bigl(\vlambda, \vtheta^\star(\vlambda)\bigr) \\
\st \quad \vtheta^\star(\vlambda)
&= \arg\min_{\vtheta \in \mathbb{R}^d} \mathcal{J}_{\text{in}}(\vlambda,\vtheta).
\end{align*}
Here, the outer objective $\outerf$ depends on the inner problem's optimal solution $\vtheta^*$. In machine learning, the inner problem typically corresponds to training a neural network $\vf(\vx;\vtheta)$ parameterized by $\vtheta$, which processes an input $\vx$ into a prediction $\vf(\vx, \vtheta) \in \sR^C$.
The prediction is compared to the ground truth $\vy$ with a criterion function $c: \vf, \vy \mapsto c(\vf, \vy) \in \sR$. 
Finally, we form the empirical risk by evaluating, then averaging, the risk over a dataset $\gD$: $\gL_{\gD}(\vtheta) = \nicefrac{1}{N} \sum_n c(\vf(\vx_n, \vtheta), \vy_n) \coloneqq \nicefrac{1}{N} \sum_n \ell_n(\vtheta)$ with the per-sample loss $\ell_n = c_n \circ \vf_n$ with $c_n = c|_{\vy = \vy_n }, \vf_n = \vf|_{\vx = \vx_n}$.
One simple example is $\mathcal{J}_{\text{in}}(\vlambda, \vtheta) = \gL_{\gD}(\vtheta) = \nicefrac{1}{N}\sum_n \ell_n (\vtheta)$, where we may identify $\vlambda = \{(\vx_n, \vy_n)\}$.

\paragraph{Hypergradient.} To optimize the outer problem, we require the gradient of $\Phi(\vlambda)$ with respect to $\vlambda$. By the implicit function theorem \citep{steven2002implicit}, the hypergradient is:
\begin{equation}
  \label{eq:hypergradient}
  \!\!\nabla \Phi(\vlambda) = \nabla_1 \mathcal{J}_{\text{out}}(\vlambda, \vtheta^\star(\vlambda)) - \nabla_{12}^2 \mathcal{J}_{\text{in}}(\vlambda, \vtheta^\star(\vlambda))\vv^\star.
\end{equation}

Here $\nabla_1$ denotes the gradient w.r.t the outer variable $\vlambda$, and $\nabla_2$ denotes the gradient w.r.t the inner variable $\vtheta$.
The subscript $_{12}$ indicates mixed second derivatives, while $_{22}$ denotes the Hessian w.r.t $\vtheta$ and $\vv^\star$ is the solution of the linear system 
\begin{align}
\label{eq:IFG-IHVP}
\nabla_{22}^2 \mathcal{J}_{\text{in}}(\vlambda, \vtheta^\star(\vlambda))\vv = \nabla_2 \mathcal{J}_{\text{out}}(\vlambda, \vtheta^\star(\vlambda)).
\end{align}
In practice, $\vv^\star$ corresponds to an IHVP, which is the main bottleneck in IFT-based bilevel optimization methods. The IFT expression above, \Cref{eq:IFG-IHVP}, holds under standard smoothness and non-degeneracy assumptions: the inner objective must be twice continuously differentiable, and its Hessian at the optimum $\vtheta^\star(\vlambda)$ must be non-singular, which implies local strong convexity \citep{steven2002implicit}. In neural networks, however, the inner optimization is rarely carried to exact optimality because of the cost, so IFT-based hypergradients are usually computed approximately at the current iterate rather than at the true inner solution.

\paragraph{IHVP approximations.}
Let $\mH \coloneqq \nabla_{22}^2 \mathcal{J}_{\text{in}}(\vlambda, \vtheta^\star(\vlambda))$ denote the inner problem's Hessian at $\vtheta^\star$, and let $\vb \coloneqq \nabla_2 \mathcal{J}_{\text{out}}(\vlambda, \vtheta^\star(\vlambda))$.
Computing the hypergradient via \Cref{eq:hypergradient} requires solving the linear system $\mH \vv = \vb$, whose exact solution corresponds to the IHVP $\vv = \mH^{-1}\vb$.
Since explicitly forming or inverting $\mH$ is infeasible in high dimensions, iterative methods are commonly used.
They iteratively solve the linear system and require one Hessian-vector product (HVP) per iteration, but their convergence is guaranteed only when $\mH$ is positive semi-definite.
An alternative view is to solve the quadratic problem
\begin{equation*}
    \vv^* = \argmin_{\vv} \frac{1}{2}\vv^\top \mH \vv - \vb^\top \vv\,,
\end{equation*}
on which iterative solvers such as conjugate gradient (CG) or gradient descent can be applied; this formulation underlies algorithms such as AmIGO \citep{arbel2022amortized, dagreou2022framework}. Other works use the Neumann series
\begin{equation*}
\mH^{-1} = \sum_{k=0}^\infty (\mI - \mH)^k\,,
\end{equation*}
and truncate it after $K$ terms to obtain an approximate IHVP with $K$ HVPs \citep{chen2021closing, ji_bilevel_2021}, though accuracy deteriorates under ill-conditioning.
Finally, one-step unrolling methods such as DARTS \citep{liu2018darts} and SAMA \citep{choe2023making} approximate $\mH$ by the identity (or equivalently $K=0$), yielding $\vv \approx \vb$, ignoring curvature in exchange for efficiency.


\paragraph{Hessian, GGN, and Fisher.}
Consider the inner loss $\gJ_{\text{in}}(\vlambda, \vtheta) = \gL_{\gD}(\vtheta)$.
To calculate the Hessian, applying the chain rule to $\ell_n$, the empirical risk's Hessian decomposes into the generalized Gauss-Newton (GGN) matrix $\mG(\vtheta)$ \citep{schraudolph2002fast} and a residual term $\mR(\vtheta)$:
\begin{equation*}
\begin{split}
  \underbrace{\nabla_{\vtheta}^2 \gL_{\gD}(\vtheta)}_{\coloneqq \mH(\vtheta)}
  &=
  \underbrace{
  \frac{1}{N} \sum_n [\mJ_{\vtheta}\vf_n]^{\top} (\nabla_{\vf}^2 c_n) [\mJ_{\vtheta}\vf_n]
  }_{\coloneqq \mG(\vtheta)}
   \\
  &+\underbrace{
  \frac{1}{N} \sum_n
  \sum_{i=1}^C [\nabla_{\vf} c_n]_i \nabla_{\vtheta}^2 [\vf_n(\vtheta)]_i
  }_{\coloneqq \mR(\vtheta)}\,.
\end{split}
\end{equation*}
Here, $\mJ_{\vtheta} \vf_n \in \sR^{C \times d}$ is the Jacobian containing the partial derivatives of $\vf_n$ \wrt $\vtheta$.
The GGN is often used instead of the Hessian as it is positive semi-definite. Moreover, for common losses such as mean-squared error and softmax cross-entropy, the GGN coincides with the Fisher information matrix \citep[FIM,][]{martens2020new}
\begin{equation*}
\scalebox{0.95}{$
  \!\mF(\vtheta)
  \!=\! \E_{\vx \sim p_{\text{data}}, \vy \sim p_\vtheta(\cdot|\vx)}
  \big[ \nabla_{\vtheta} \log p_\vtheta(\vy|\vx) \,
        \nabla_{\vtheta} \log p_\vtheta(\vy|\vx)^\top \big]\!
$}
\end{equation*}
where $p_{\text{data}}(\vx)$ is the input distribution and $p_\theta(\vy|\vx)$ is the model’s predictive distribution. The Fisher view underlies natural gradient descent \citep{amari1998natural} and motivates KFAC as a preconditioner for neural network optimization \citep{martens2015optimizing}. While both the Hessian and GGN are prohibitively expensive to store, automatic differentiation allows cheap evaluation of matrix-vector products at the same complexity as computing the gradient \citep{pearlmutter1994fast,schraudolph2002fast,dagreou2024how}. However, this computation is still expensive for BO as it requires applying an inverse at every outer iteration.


\section{Method}
Hypergradient computation in BO requires IHVPs, which are expensive or unstable with iterative solvers.
We replace them with IKVPs, enabled by a layer-wise Kronecker-factored approximation of the GGN/Fisher.
\Cref{sec:kfac} recalls KFAC, \Cref{sec:replace_IHVP} shows how to integrate IKVPs into BO, and \Cref{sec:diagonal} provides a diagnostic study that motivates and illustrates its robustness under ill-conditioning.

\subsection{KFAC: a Layer-wise Curvature Approximation}
\label{sec:kfac}
The Hessian $\mH$ or GGN $\mG$ is $d \times d$, which is prohibitively large for modern architectures.
KFAC reduces this cost by adopting a layer-wise approximation, treating each layer independently and ignoring cross-layer interactions.
This yields a block-diagonal structure, where each block is further approximated with a Kronecker product.
This structure is particularly attractive for BO, since hypergradients require repeated curvature inverses: KFAC replaces expensive iterative solvers with direct structured inverses, enabling curvature information at low cost.

\paragraph{KFAC for a single fully-connected layer.}
Here we consider the case for a fully-connected layer inside a neural net. The layer's weights are $\mW \in \sR^{d_1 \times d_2}$ (we omit biases for simplicity) and it processes an input $\va_n \in \sR^{d_2}$ into an output $\vz_n = \mW \va_n \in \sR^{d_1}$.
Further, let $\flatten(\mW) \in \sR^{d_1 d_2}$ denote the flattened weights from stacking the rows into a vector.
We can then write down the GGN matrix $\mG(\flatten\mW)$ \wrt to the layer's parameters analytically.
To do so, we use the chain rule ($\mJ_{\flatten\mW} \vf_n = (\mJ_{\vz_n} \vf_n) (\mJ_{\flatten \mW} \vz_n) $) with analytical output-parameter Jacobian $\mJ_{\flatten \mW} \vz_n = \mI_{d_1} \otimes \va_n^{\top}$ \citep{dangel2020modular}, yielding that each datum contributes a single Kronecker product:
\begin{align*}
  \mG(\flatten \mW)
  &= \frac{1}{N} \sum_n
    (\mJ_{\vz_n} \vf_n)^{\top}
    (\nabla_{\vf}^2 c_n)
    (\mJ_{\vz_n} \vf_n)
    \otimes \va_n \va_n^{\top} \notag \\
  &= \hat\E\!\left[
    (\mJ_{\vz} \vf)^{\top}
    (\nabla_{\vf}^2 c)
    (\mJ_{\vz} \vf)
    \otimes \va \va^{\top}
  \right]
\end{align*}
where we use the notation $\hat\E[\bullet] = \nicefrac{1}{N}\sum_n \bullet_n$ to indicate the expectation over the dataset.
To obtain a single Kronecker product, KFAC makes the expectation approximation $\hat\E[\bullet \otimes \star] \approx \hat\E[\bullet] \otimes \hat\E[\star]$.
In summary, this yields the following approximation of the GGN:
\begin{align*}
    \mG(\flatten \mW) &\approx \mB_{\text{KFAC}} \otimes \mA_{\text{KFAC}}
    \shortintertext{where}
\mB_{\text{KFAC}} &=  \frac{1}{N} \sum_n (\mJ_{\vz_n} \vf_n)^{\top}(\nabla_{\vf}^2 c_n)(\mJ_{\vz_n} \vf_n), 
\\
\mA_{\text{KFAC}} &= \frac{1}{N} \sum_n \va_n \va_n^{\top}. 
\end{align*}
This structural approximation reduces the storage cost of  $\mG(\flatten \mW) \in \sR^{d_1 d_2 \times d_1 d_2}$ to the much smaller Kronecker factors $\mA_{\text{KFAC}} \in \sR^{d_2 \times d_2}$ and $\mB_{\text{KFAC}} \in \sR^{d_1 \times d_1}$ for each layer. Inverting KFAC resorts to inverting the two Kronecker factors, since $(\mB \otimes \mA)^{-1} = \mB^{-1} \otimes \mA^{-1}$. For a network with multiple layers, KFAC applies this construction layerwise and combines the resulting factors into a block-diagonal approximation of the full network's curvature.

\paragraph{Computational details.}
The (uncentered) covariance of the layer inputs, $\mA_{\text{KFAC}}$, is easy to compute in a forward pass.
For the first factor $\mB_{\text{KFAC}}$, there exist multiple variants that differ in computational cost (see \citealt{dangel2025kfac} for a detailed introduction).
The most expensive one \citep{botev2017practical} employs a symmetric factorization of the criterion's Hessian into $\smash{\nabla_{\vf}^2} c_n = \nicefrac{1}{C} \smash{\sum_{i=1}^C} \vs_{n,i} \vs_{n,i}^{\top}$, then computes the covariance of the pseudo-gradients $\smash{\tilde{\vg}_{n, i} \coloneqq (\mJ_{\vz_n} \vf_n)^{\top} \vs_{n,i}}$, \ie $\mB_{\text{KFAC}} = \nicefrac{1}{NC} \sum_n \sum_i \tilde{\vg}_{n,i} \tilde{\vg}_{n,i}^{\top}$.
This incurs $C$ backpropagations per datum.
We will follow the original approach of \citet{martens2015optimizing}, which introduces a randomization via $p(\vs_n)$ such that $\smash{\nabla_{\vf}^2 c_n = \E_{\vs_{n,i} \sim p(\vs_n)}[ \vs_{n,i} \vs_{n,i}^{\top}]}$, then approximates the expectation with a Monte-Carlo estimate involving $M$ samples, \ie $\nabla_{\vf}^2 c_n \approx \nicefrac{1}{M} \sum_i \vs_{n,i} \vs_{n,i}^{\top}$ where $\vs_{n,i } \smash{\stackrel{\text{\iid}}{\sim}} p(\vs_n)$.
This reduces the number of backpropagations from $C$ to $M$, and we will use $M=1$ as is common practice.
The distribution $p(\vs_n)$ is determined by the probabilistic interpretation of the loss function (e.g., Gaussian for MSE; the explicit forms are given in \Cref{app:hessian_sampling}).



We also adopt the empirical variant of KFAC (KFAC EMP), which constructs $\mB_{\text{KFAC}}$ from empirical, rather than sampled, gradients. Its main computational advantage is that it reuses the standard backward pass already needed for gradient computation, while Monte-Carlo GGN variants typically require an additional backward pass with sampled curvature probes. The empirical $\mB_{\text{KFAC}}$ is estimated as
\begin{equation*}
    \mB_{\text{KFAC}} = \tfrac{1}{N}\sum\nolimits_{n} \vg_n \vg_n^\top, 
    \qquad 
    \vg_n = \nabla_{\vz_n} \ell_n(\vtheta),
\end{equation*}
with $\vg_n$ denoting the per-example gradient of the loss with respect to the layer pre-activations. 
The remaining components of KFAC, namely the Kronecker factorization, damping, and matrix inversion, are unchanged. 
Empirical Fisher approximations of this form are widely used in practical second-order optimization \citep{george2018fast, lin2023structured, zhang2022scalable}. 
We compare KFAC EMP with standard KFAC in \Cref{sec:bert}, and use KFAC EMP in all experiments in \Cref{sec:broader application} due to the size of problems.

\paragraph{KFAC approximation accuracy.} For linear networks with square loss, the Hessian, GGN, and KFAC are identical \citep{petersen2023isaac}, yielding exact natural-gradient updates and  IHVPs. This exactness further extends to deep linear networks: despite nonlinearity in parameters, KFAC recovers the exact natural gradient up to a constant factor, and is equivalent to the block-diagonal GGN in this setting \citep{bernacchia2018exact}. Beyond these regimes, KFAC provides a structured approximation to the GGN for deep nonlinear networks, preserving curvature information at layerwise granularity while remaining computattionally tractable. Moreover, in the infinite-width in the NTK regime \citep{jacot2018neural}, approximate natural gradient descent via KFAC achieves the same convergence rate as exact gradient methods \citep{karakida2020understanding}. Our work focuses on practical large-scale regimes where exact solvers are infeasible, and evaluates how effectively this approximation supports hypergradient computation in BO.


\subsection{Replacing IHVP with IKVP}
\label{sec:replace_IHVP}
We focus on BO problems where hypergradients can be expressed, under the implicit function theorem, as in \eqref{eq:hypergradient}. Instead of approximating IHVP $\mH^{-1}\vb$, we follow the common practice of approximating $\mH$ with the GGN matrix $\mG$, and approximating
$\mG^{-1}\vb$ instead.

To scale this computation, we further approximate $\mG$ with a Kronecker-factored block-diagonal estimate $\hat{\mG}$ obtained via KFAC. Suppose we would like to apply $\hat{\mG}^{-1}\vv$ for some parameter vector $\vv$. Since $\hat{\mG}$ is block diagonal, for simplicity, we can just consider a single layer case: for $\mW \in \mathbb{R}^{d_1 \times d_2}$, let $\mA_{\text{KFAC}} \in \sR^{d_2\times d_2}$ denote the uncentered covariance matrix of the activation and $\mB_{\text{KFAC}} \in \sR^{d_1\times d_1}$ denote the gradient covariances. By construction, the KFAC block takes the form $\hat{\mG} = \mB_{\text{KFAC}} \otimes \mA_{\text{KFAC}}$. Given a reshaped vector $\vv = \operatorname{vec}(\mV)$, where $\mV$ matches the shape of the layer's weights, the IKVP follows from Kronecker identities as
\begin{equation*}
    \hat{\mG}^{-1} \vv = (\mB_{\text{KFAC}}^{-1}
    \otimes \mA_{\text{KFAC}}^{-1}) \vv = \operatorname{vec} \big( \mB_{\text{KFAC}}^{-1} \mV \mA_{\text{KFAC}}^{-1} \big).
\end{equation*}
Thus, instead of applying the inverse of a full $d_1d_2 \times d_1d_2$ matrix, IKVP reduces to two smaller inversions and a sequence of matrix multiplications.

\begin{figure*}[!t]
    \centering
    \includegraphics[width=\linewidth]{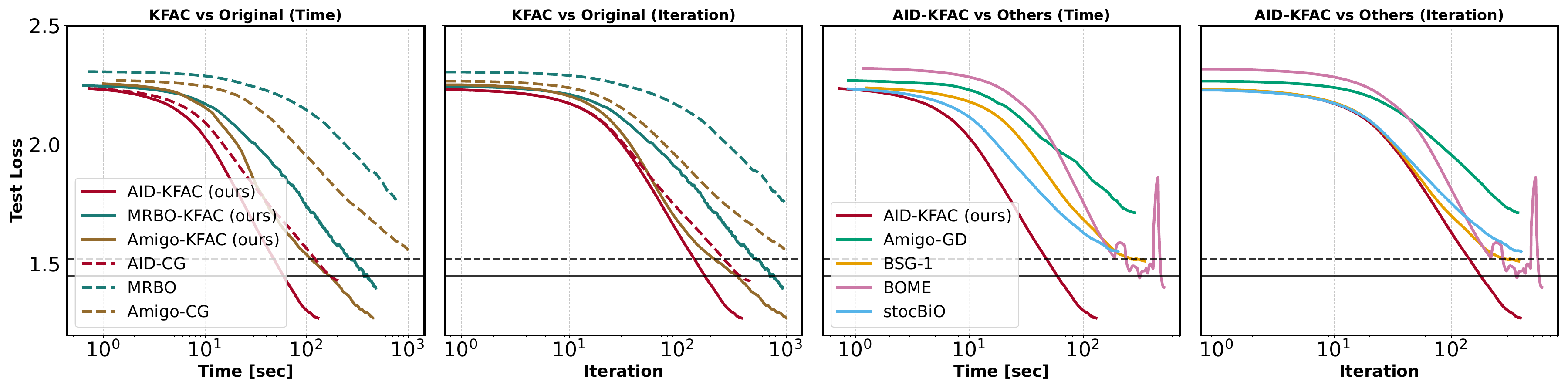}
    \vspace{-4ex}
    \caption{Results on data hypercleaning. \textbf{Left two panels:} existing BO algorithms improve when their IHVP solvers are replaced with IKVPs. \textbf{Right two panels:} KFAC-based methods compared against other non-IFT baselines, showing faster convergence in both time and iterations. Curves are truncated at the lowest test loss reached. Solid black line is trained on the validation dataset, while dashed blackline is trained on the validation dataset and the noisy dataset together.}
    \label{fig:ResNet18 Convergence}
\end{figure*}

\paragraph{Computational Cost of IKVP.} For each layer, inverting $\mA_{\text{KFAC}}$ and $\mB_{\text{KFAC}}$ requires $O(d_1^3 + d_2^3)$ operations. Once inverted, applying an IKVP to a vector costs $O(d_1^2d_2 + d_1d_2^2)$, corresponding to two matrix-matrix multiplications. Performing forward and backward passes is $O(B d_1 d_2)$, where $B$ is the batch size, so this operation has similar complexity to backpropagation when $d_1$ and $d_2$ are similar to $B$.

\paragraph{Damping.} Curvature approximations are known to be singular for neural nets \citep{sagun2016eigenvalues}. To stabilize inversion, we use the factored Tikhonov damping, which adds scaled identity to each Kronecker factor, rather than to the full block, which preserves efficiency while improving conditioning. In our implementation, we adopt the \emph{heuristic damping} strategy from \citet{martens2015optimizing}, where the relative scaling between the activation covariance $\mA_{\text{KFAC}}$ and the gradient covariance $\mB_{\text{KFAC}}$ is controlled by a parameter $\pi$ derived from their traces. Concretely, the damping applied to $\mA_{\text{KFAC}}$ and $\mB_{\text{KFAC}}$ is
\begin{equation*}
\mA_{\text{KFAC}} \rightarrow \mA_{\text{KFAC}} + \pi \sqrt{\lambda} \mI, \quad \mB_{\text{KFAC}} \rightarrow \mB_{\text{KFAC}} + \tfrac{1}{\pi} \sqrt{\lambda} \mI,
\end{equation*}
with $\pi = \sqrt{ \nicefrac{d_2\text{Tr}(\mA_{\text{KFAC}})}{d_1\text{Tr}(\mB_{\text{KFAC}})} }$ balancing the two factors. Heuristic damping has been shown to perform more robustly than naive Tikhonov damping \citep{martens2015optimizing}, while adding negligible computational overhead.

there exist many public source implementations of KFAC \citep{dangel2020backpack,osawa2023asdl,grosse2023studying,botev2022kfac-jax}.
We use the \texttt{curvlinops} package \citep{dangel2025position}, which provides easy access to VPs products with KFAC and its damped inverse.

\subsection{Diagnostic Study of IHVP Approximations}
\label{sec:diagonal}

A key practical advantage of KFAC arises from its robustness to ill-conditioning in the inner-level curvature.
To isolate this effect, we consider a controlled setting where the exact inverse Hessian is available and
compare different IHVP approximations under increasing dimensionality.

We study linear regression with square loss, where both design matrix $\mX \in \mathbb{R}^{d \times N}$ and targets $\vy \in \mathbb{R}^N$ are generated with \iid standard Gaussian entries. We fit $\vw \in \mathbb{R}^{d}$ using the square loss. The Hessian is known in closed form and only depends on the data matrix:
$$\mH = \frac{1}{N} \mX \mX^\top.$$
We vary the parameter dimension $d$ while keeping the dataset size $N$ to 100. In this setting, KFAC is an exact approximation of the Hessian \citep{petersen2023isaac}. As $d$ grows relative to $N$, the Hessian becomes increasingly ill-conditioned and eventually rank-deficient, a regime commonly encountered in over-parameterized models.

We compare Monte-Carlo KFAC against truncated Neumann series
and CG under fixed computational budgets.
Accuracy is measured by the relative operator-norm error
$$\min_{\alpha}\|\alpha (\widehat{\mH} + \lambda\mI)^{-1} - (\mH + \lambda \mI)^{-1}\|_2 / \|(\mH + \lambda \mI)^{-1}\|_2,$$ 
where $\alpha$ is searched to account for the step-size in training and $\lambda = 10^{-5}$ is the damping factor.

\Cref{tab:toy-convex} highlights two consistent trends. First, while CG and Neumann solvers can achieve high accuracy at small $d$ given sufficient iterations,
their error degrades rapidly as $d$ increases under fixed iteration budgets.
This behavior is expected, as the convergence rate of iterative solvers depends on the Hessian's condition number \citep{bakushinsky2004iterative}.
Second, KFAC maintains stable accuracy as $d$ increases, achieving orders-of-magnitude lower error
than budget-limited CG and Neumann solvers.

\begin{table}[t]
\centering
\small
\setlength{\tabcolsep}{4pt}
\caption{Relative error for different IHVP approximations as the parameter dimension $d$ increases. Neu-$K$ denotes a truncated Neumann series with $K$ terms. CG-$T$ denotes conjugate gradient with $T$ iterations. Identity corresponds to the identity approximation used in one-step unrolling.}
\label{tab:toy-convex}
\setlength{\tabcolsep}{4pt}
\small
\begin{tabular}{l S S S}
\toprule
Method & {$d=10$} & {$d=100$} & {$d=500$} \\
\midrule
KFAC & \num{7.62e-3} & \num{7.62e-3} & \num{7.62e-3} \\
\midrule
Neu-3  & \num{1.12e-1} & \num{2.45e-1} & \num{8.01e-1} \\
Neu-20 & \num{2.22e-2} & \num{4.46e-2} & \num{2.41e-1} \\
Neu-50 & \num{1.01e-4} & \num{5.62e-3} & \num{1.39e-1} \\
\midrule
CG-3       & \num{1.05e-3} & \num{7.49e-2} & \num{3.02e-1} \\
CG-5      & \num{6.31e-4} & \num{7.74e-3} & \num{1.96e-1} \\
CG-10      & \num{6.31e-4} & \num{8.68e-4} & \num{7.75e-2} \\
\midrule
Identity ($\mH=\mI$) & \num{1.69e-1} & \num{3.91e-1} & \num{4.93e-1} \\
\bottomrule
\end{tabular}
\end{table}


\section{Enhancing BO Algorithms with KFAC}
\label{sec:enhance_bo}

Here, we demonstrate that our KFAC-based IKVP solver can serve as a direct, drop-in replacement for the expensive IHVP solver used in existing IFT-based BO algorithms. Our method yields faster convergence (in wall-clock time) and improved performance, especially at larger scales.

\subsection{Data Hypercleaning with ResNet-18}
\label{sec:exp-overview}

We first test our method on data hypercleaning, a standard and widely used benchmark for evaluating BO convergence \citep{ye_bome_2022,ji_bilevel_2021,dagreou2022framework}. Our goal is to isolate the benefit of replacing the IHVP subroutine in IFT-based methods with our IKVP solver.

Given a corrupted training set \(\dataset{tr} = \{(\vx_n, y_n)\}_{n=1}^N\) (random label noise) and a clean validation set $\dataset{val}$, we learn per-example weights $\vlambda \in [0, 1]^N$ by the bilevel objective:
\begin{equation}
\label{eq:data-hypercleaning}
\begin{aligned}
\min_{\vlambda, \vtheta} \quad & \gL^{\text{val}}(\vtheta^\star), \\
\text{s.t.} \quad & \vtheta^\star \in \argmin\nolimits_{\vtheta}
\left\{ \gL^{\text{tr}}(\vtheta, \vlambda) + \alpha\|\vtheta\|^2 \right\},
\end{aligned}
\end{equation}
where \(\gL^{\text{val}}(\vtheta^\star)\) is the cross-entropy loss evaluated on the validation set.
\(\gL^{\text{tr}}(\vtheta, \vlambda)\) is the weighted training loss, defined as: $\gL^{\text{tr}}(\vtheta, \vlambda) = \sum_{n=1}^N \sigma(\lambda_n) \ell_n(\vx_n, y_n, \vtheta)$,
where \(\sigma(\lambda_n) = \text{Clip}(\lambda_n, [0, 1])\) ensures that \(\lambda_n \in [0, 1]\).
The training loss \(\ell(\vx_n, y_n, \vtheta)\) is also cross-entropy,
and \(\alpha \|\vtheta\|^2\) adds an \(L_2\) regularization to prevent overfitting. Since the GGN scales with the sample weight $\sigma_n$, in the Monte Carlo approximation, we rescale each pseudo-loss by $\sqrt{\sigma_n}$ for unbiased curvature estimation (details in \Cref{app:weighted_fisher}).




Previous works focusing on theoretical guarantees typically use MNIST with a linear model, far from practical scenarios.
We instead evaluate on CIFAR-10 with a three-layer CNN and ResNet-18 to stress scalability.
We adapt AmIGO \citep{arbel2022amortized}, MRBO \citep{ji_bilevel_2021}, and AID-CG \citep{grazzi2020iteration}, all IFT-based methods that approximate the IHVP. We replace their solver with IKVPs and compare against other methods including the IFT-based stocBiO \citep{ji_bilevel_2021}, and first-order methods BOME \citep{ye_bome_2022}, BSG-1 \citep{giovannelli2021bilevel}, stocBiO \citep{ji_bilevel_2021}. We also replace the Hessian with the GGN and apply Neumann and CG. For fairness, we use CG with 3 iterations and Neumann with 10 terms, which strikes a balance between runtime and accuracy. We run every method for fixed iterations to ensure they reach the minimum validation loss. Hyperparameters, including inner/outer learning rates, momentum, and damping values, were selected via grid search on a dedicated validation set.

\Cref{fig:ResNet18 Convergence} highlights our main experimental finding: replacing IHVPs with IKVPs yields clear efficiency gains. The plots, which track the lowest test loss achieved within 1000 iterations, show that KFAC adapted methods (AID-KFAC, MRBO-KFAC, Amigo-KFAC) converge faster in wall-clock time and improve iteration-wise performance compared to their original IHVP-based counterparts. In \Cref{fig:intro}, we vary the number of CG iterations and Neumann terms within the AID algorithm.
While the GGN is slightly more efficient than the Hessian, both remain weaker than KFAC in terms of the balance between accuracy and runtime. We detail the experimental results for a three-layer CNN and a linear model in \Cref{app:Data Hypercleaning}, where the performance gap between KFAC and the baseline solvers is less pronounced on these simpler architectures than on the more complex ResNet-18.

\subsection{KFAC performs better across different batch sizes}

Using a smaller batch size is a common technique to reduce computational cost. However, small batches introduce noise, which can destabilize curvature estimates and harm solvers like CG that are sensitive to ill-conditioning \citep{granziol_learning_2021}. We want to test the robustness of KFAC compared to CG as the batch size is varied.

\begin{figure}[!t]
\centering
\includegraphics[width=\linewidth]{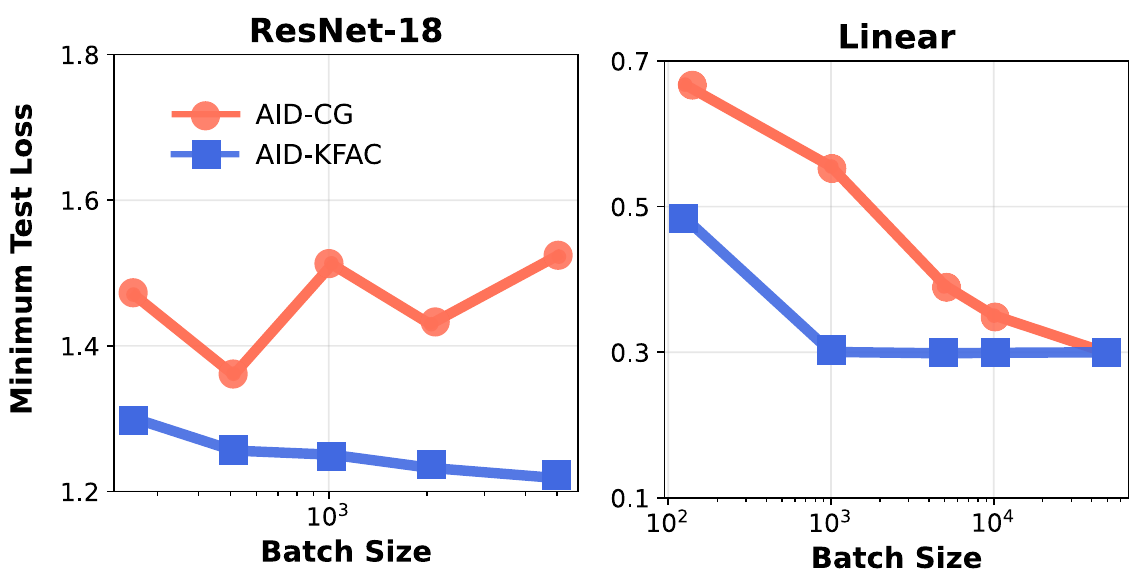}
\vspace{-4ex}
\caption{Batch size effects on data hypercleaning. KFAC consistently achieves lower minimum test loss than CG across models and batch sizes. For linear model, CG will converge early to a worse point than KFAC.} 
\label{fig:batch size study}
\end{figure}

We study this under the data hypercleaning setting with a linear model and ResNet-18. We solve the BO problem for 1000 iterations across varying batch sizes and record the minimum test loss and time taken to reach that loss (\Cref{fig:batch size study}).
For the linear model, full convergenc is achievable: both methods perform poorly with small batches but improve as batch size grows, eventually closing the gap between KFAC and CG, likely due to instability in hypergradient estimates under small batches.
We give more details in \Cref{app:batch size}. For ResNet-18, where full convergence is unattainable, no monotonic improvement is seen, yet KFAC consistently outperforms CG at all batch sizes.

\subsection{EKFAC and Amortization}
In second-order optimization, it is common to amortize KFAC's curvature factors across iterations to save cost. 
Also, Eigenvalue-Corrected KFAC (EKFAC, \citealt{george2018fast}) is a popular refinement that improves KFAC's accuracy.
We study both effects on ResNet-18 data hypercleaning by comparing KFAC to EKFAC under factor-update intervals $\tau \in \{1, 5, 10, 50\}$: 
gradients are recomputed every step but curvature is frozen within an interval.

We summarize the trade-off between computational efficiency and model performance in \Cref{tab:kfac_ekfac_side_by_side}. As expected, amortizing curvature computation drastically reduces the per-iteration runtime. Specifically, increasing the update interval from $\tau=1$ to $\tau=10$ reduces KFAC's runtime by approximately $62\%$ ($0.69$s to $0.26$s). Consistent with our analysis of the computational overhead, EKFAC is significantly slower than KFAC at low intervals due to the costly eigendecomposition, though this gap narrows at larger $\tau$. Crucially, the test loss remains robust under amortization; KFAC achieves its lowest test loss at $\tau=5$ ($1.33$), suggesting that moderate amortization ($\tau=5$ or $10$) yields an optimal balance between efficiency and accuracy. Besides EKFAC and amortization, we also ablate the commonly used EMA technique \citep{martens2015optimizing} in \Cref{app:EMA}, and we show that EMA can help stabilize the convergence for the linear model under small batch size.

\begin{table}[t]
\centering
\label{tab:kfac_ekfac_side_by_side}
\begin{tabular}{c|cc|cc}
\toprule
\multirow{2}{*}{\textbf{$\tau$}} & 
\multicolumn{2}{c|}{\textbf{Best Test Loss}} & 
\multicolumn{2}{c}{\textbf{Avg. iter Time (s)}} \\
 & \textbf{KFAC} & \textbf{EKFAC} & \textbf{KFAC} & \textbf{EKFAC} \\
\midrule
1   & 1.24 & 1.21 & 0.69 & 1.83 \\
5   & 1.33 & 1.24 & 0.28 & 0.62 \\
10  & 1.35 & 1.30 & 0.26 & 0.36 \\
50  & 1.36 & 1.33 & 0.22 & 0.28 \\
\bottomrule
\end{tabular}
\caption{Best test loss and average per-iteration runtime for KFAC and EKFAC with varying update intervals ($\tau$) on ResNet-18. Amortization significantly reduces computational cost while maintaining competitive performance.}
\end{table}

\begin{table*}[!t]
\centering
\caption{ResNet-32 accuracy (\%) on long-tailed CIFAR-10/100 with different imbalanced factors (IFs), best is in \textbf{bold}.}
\label{tab:ltcifar}
\setlength{\tabcolsep}{10pt}
\renewcommand{\arraystretch}{1.0}
\resizebox{\textwidth}{!}{
\begin{tabular}{lccccc|ccccc}
\toprule
Dataset & \multicolumn{5}{c|}{CIFAR-10} & \multicolumn{5}{c}{CIFAR-100} \\
IF & 200 & 100 & 50 & 20 & 10 & 200 & 100 & 50 & 20 & 10 \\
\midrule
Base Model        & 65.3 & 70.0 & 74.2 & 81.8 & 86.1 & 34.4 & 38.5 & 43.4 & 50.7 & 55.5 \\
Focal Loss        & 65.2 & 69.9 & 74.3 & 81.5 & 85.9 & 34.9 & 38.4 & 43.7 & 51.8 & 55.6 \\
MWN + $T_1$-$T_2$ & 66.3 & 72.8 & 78.6 & 84.8 & 87.5 & \textbf{36.0} & 40.2 & 46.0 & 52.0 & 58.6 \\
MWN + SAMA        & 66.3 & 72.9 & \textbf{79.5} & 85.4 & 87.6 & 35.8 & 40.0 & 45.7 & 52.5 & 59.7 \\
MWN + KFAC (ours) & \textbf{67.0} & \textbf{73.1} & 78.9 & \textbf{85.6} & \textbf{88.4} & 35.6 & \textbf{40.9} & \textbf{46.2} & \textbf{53.2} & \textbf{60.8} \\
\bottomrule
\end{tabular}
}
\end{table*}

\section{Curvature Matters at Scale}
\label{sec:bert}

\begin{figure}[!t]
    \centering
    \includegraphics[width=1\linewidth]{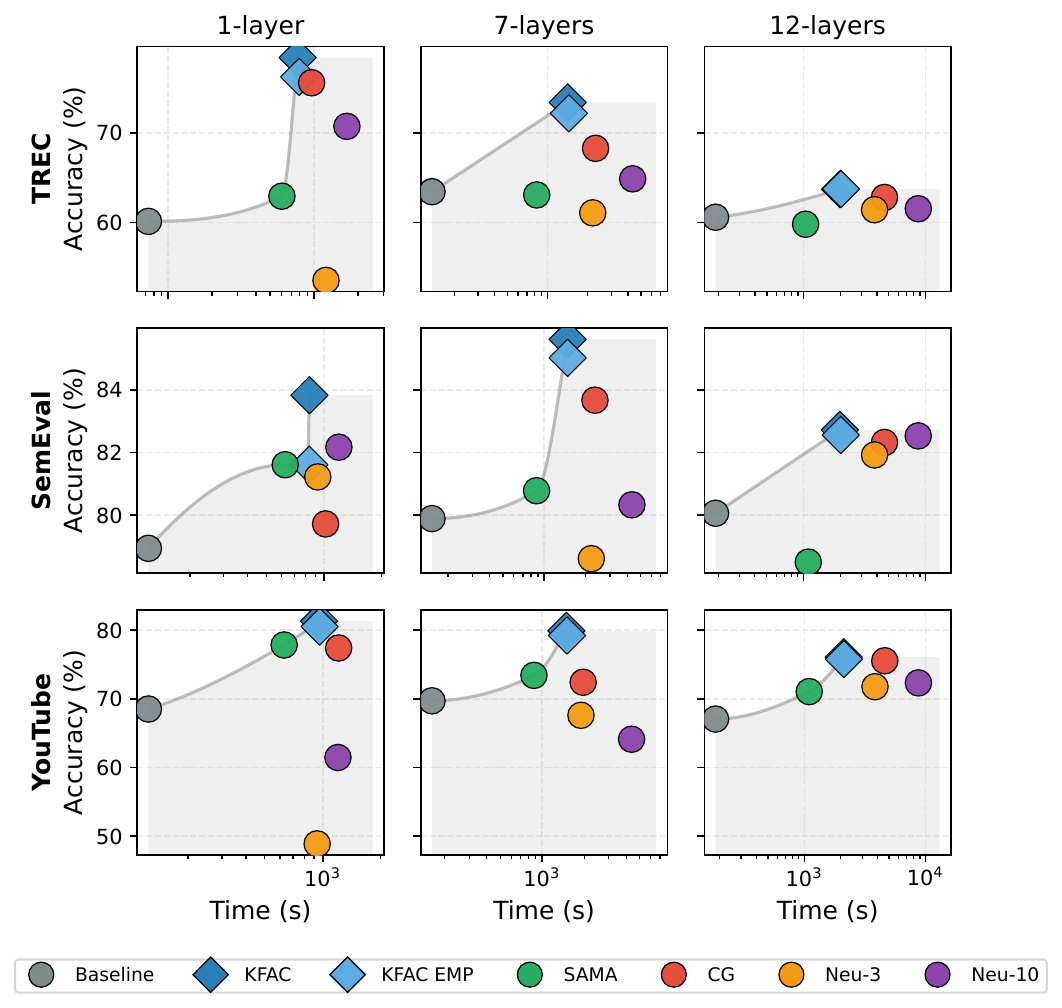}
    \vspace{-4ex}
    \caption{Results on BERT data hypercleaning. We report test accuracy (y-axis) versus total training time (x-axis) on three datasets when fine-tuning 1, 7, 12 encoder layers (columns). Each marker corresponds to one hypergradient solver. The gray region highlights the Pareto frontier.}
    \label{fig:Pareto}
\end{figure}

Scaling BO to transformers poses unique challenges: curvature matrices such as the Hessian or GGN become prohibitively large, making direct solvers infeasible. Scalable baselines like SAMA \citep{choe2023making} sidestep this by ignoring curvature, achieving low cost but sacrificing accuracy and stability. In this section, we address two critical questions for scaling BO to massive models like BERT:
\begin{enumerate}[leftmargin=*]
    \item Is curvature still beneficial at this scale, compared to gradient-only approximations (like SAMA)?
    \item Can KFAC provide curvature information scalably and efficiently, without becoming prohibitively expensive?
\end{enumerate} 

To answer this, we design an experiment on three text classification datasets from the WRENCH benchmark \citep{zhang2wrench}. We use the same data hypercleaning formulation as \eqref{eq:data-hypercleaning}, but remove the $L_2$ regularization (for a large-scale model like BERT, this sums the squares of over 100 million parameters, which can become numerically unstable or dominate the inner task loss). To systematically vary the inner problem size and to make Hessian-based baselines feasible, we freeze different numbers of encoder layers in BERT and gradually unfreeze them starting from the last layer. This avoids out-of-memory issues for Neumann and CG and allows us to study solver behaviour as more layers are trained, from only the last layer ($\approx$6\% of parameters) up to the full 12-layer model. 
We compare SAMA, Neumann with 3 or 10 terms, CG, and two variants of our approach, KFAC and KFAC EMP, trained for 1000 bilevel iterations under the AID-style double-loop framework. We report validation accuracy, runtime, and memory footprint in \Cref{fig:intro,fig:Pareto}.

Across all tuning depths, KFAC consistently attains the best accuracy, showing that curvature information remains beneficial at BERT scale. At the same time, its runtime stays in the same overall range as competing baselines, while its memory use remains manageable. As more layers are unfrozen, KFAC continues to perform reliably, whereas CG and Neumann become less practical due to memory limits or degraded performance. Interestingly, the layer-wise trend suggests that restricting the inner problem to a single layer can yield higher accuracy than full fine-tuning, meaning that the regularization provided by a smaller parameter space can improve the stability of hypergradient estimation. Further details can be found in \Cref{app:bert_more_explanation}.

\section{Broader Applicability}
\label{sec:broader application}

\begin{table}[!t]
\centering
\Large
\caption{Experiment results for auxiliary learning with the continued pretraining task. Following \citet{choe2023making}, we report test micro-F1 for ChemProt and macro-F1 for other datasets; the result is averaged over 3 runs.}
\renewcommand{\arraystretch}{1.3}
\resizebox{\columnwidth}{!}{%
\begin{tabular}{lcccc}
\toprule
 & ChemProt & HyperPartisan & ACL-ARC & SciERC  \\
\midrule
Baseline & 82.54$_{\pm0.32}$ & 90.05$_{\pm1.24}$ & 67.85$_{\pm2.03}$ & 79.87$_{\pm0.59}$ \\
TARTAN-MT & 83.97$_{\pm0.25}$ & 94.64$_{\pm0.91}$ & 72.86$_{\pm3.01}$ & 80.49$_{\pm0.82}$ \\
SAMA & 83.75$_{\pm0.24}$ & \textbf{95.18}$_{\pm0.03}$ & 71.75$_{\pm1.65}$ & 81.06$_{\pm0.09}$ \\
KFAC (ours) & \textbf{85.04}$_{\pm0.23}$ & 95.13$_{\pm0.05}$ & \textbf{72.98}$_{\pm1.58}$ & \textbf{81.22}$_{\pm0.14}$ \\
\bottomrule
\end{tabular}
}
\label{tab:continued_pretraining}
\end{table}

\begin{table*}[!t]
\caption{Accuracy drop (\%) and training time of different data poisoning algorithms on LR, MLP, and CNN. Results are averaged over 5 runs. Label flipping requires no training, so time is omitted.}
\vspace{-2ex}
\label{table:data_poison}
\begin{center}
\begin{small}
\begin{sc}
\resizebox{\textwidth}{!}{%
\begin{tabular}{ccc|cc|cc|cc}
\toprule
\multirow{2}*{Model} & \multirow{2}*{Dataset} & Clean & \multicolumn{2}{c|}{TGDA Hessian $\to$ KFAC (Ours)} & \multicolumn{2}{c|}{BOME (VF)} & \multicolumn{2}{c}{Label Flip} \\
~ & ~ & Acc & Acc Drop & time & Acc Drop & time & Acc Drop & time \\
\midrule
LR & MNIST & 92.34 & $2.78_{\pm 0.04} \to \textbf{5.40}_{\pm 0.06}$ & 0.7 $\to$ 0.4 hrs & $2.22_{\pm 0.04}$ & 0.4 hrs & $1.52_{\pm 0.03}$ & -\\
MLP & MNIST & 98.02 & $1.48_{\pm 0.04} \to \textbf{3.99}_{\pm 0.05}$ & 7.3 $\to$ 3.2 hrs &$1.63_{\pm 0.05}$ & 3.0 hrs & $0.05_{\pm 0.01}$ & -\\
CNN & CIFAR-10 & 64.30 & $2.04_{\pm 0.03} \to \textbf{3.44}_{\pm 0.07}$ & 21.5 $\to$ 6.2 hrs & $2.15_{\pm 0.03}$ & 5.9 hrs & $0.17_{\pm 0.02}$ & - \\
\bottomrule
\end{tabular}%
}
\end{sc}
\end{small}
\end{center}
\vskip -0.1in
\end{table*}
Thus far, we have focused on the data hypercleaning problem, but BO underlies many other machine learning tasks. To test generality, we integrate KFAC into existing frameworks for meta-learning and AI safety. Across these domains, 
we used KFAC EMP.
This section shows our final experimental contribution: our method seamlessly applies across domains and achieves better performance. All experimental details of this section are in \Cref{app:more application details}.

\subsection{Meta Learning Applications}

Meta-learning seeks to acquire inductive biases, such as sample reweighting or task-specific initialization, that enable downstream models to adapt efficiently to user-specific objectives. These problems naturally admit a bilevel structure and form an ideal testbed for our method. We integrate KFAC into the \textit{Betty} library \citep{choe2022betty} and study two tasks: 


\paragraph{(1) Meta-Weight-Net (MWN) for Class Imbalance.}

This task addresses training under long-tailed class distributions, where some classes have far fewer examples than others, by learning a meta-network that assigns adaptive sample weights to balance training. Following \citet{shu2019meta}, we construct long-tailed CIFAR-10 and CIFAR-100 by exponentially decaying the number of samples per class. A small validation set of 10 images per class is used as meta-data. The original MWN work employed one-step unrolling ($T_1-T_2$) to approximate the bilevel objective. We compare this baseline to alternatives where the inner problem is solved with different hypergradient approximations: MWN with SAMA and MWN with our proposed KFAC approximated hypergradients. 
We also compare to another baseline trained using the focal loss \citep{LinGGHD20}.
\Cref{tab:ltcifar} summarizes the results. On CIFAR-10, KFAC achieves the best accuracy for most imbalances (e.g., 88.4\% vs.\,87.6\% for MWN+SAMA at imbalance factor 10). On CIFAR-100, the advantage is clearer: KFAC improves over SAMA by up to +1.1\% under severe imbalance. This suggests that additional curvature information stabilizes MWN training under extreme data imbalance.

\paragraph{(2) Continued Pretraining of Large  Language Models.} We next evaluate continued pretraining on four domain-specific corpora, where BO is used to balance auxiliary and end-task objectives.
Baselines include TARTAN \citep{dery2021should} and SAMA \citep{choe2023making}.
As shown in \Cref{tab:continued_pretraining}, KFAC achieves the strongest or tied-best performance on three datasets: +1.1\% on ChemProt compared to TARTAN, and +0.2–1.2\% on ACL-ARC compared to SAMA.
On HyperPartisan, all methods exceed 95\%, leaving little room for improvement.
These results indicate that KFAC can be seamlessly integrated into large-scale language model pretraining pipelines, where even small accuracy improvements are valuable.

\subsection{AI Safety Applications}
Many safety-critical machine learning problems can be seen as BO, often in the form of an attacker-defender game. The outer-level variable controls adversarial perturbations or poisoned data, while the inner-level learner attempts to minimize training loss on potentially corrupted inputs. Such problems are particularly challenging due to their adversarial nature and the need to scale to larger models like CNNs and ResNets. We evaluate KFAC on two standard testbeds: data poisoning and unlearnable examples.

\paragraph{Data Poisoning.} We 
evaluate on MNIST (LR/MLP) and CIFAR-10 (CNN) with a small poisoning ratio ($\epsilon=3\%$), comparing against TGDA \citep{lu2022indiscriminate}, BOME, and label flipping ($y \rightarrow 10 - y$). Following \citet{lu2022indiscriminate}, evaluation involves attacker pretraining, joint training, and testing via defender retraining on poisoned data. As shown in \Cref{table:data_poison}, our KFAC-based method consistently surpasses baselines, matching BOME’s efficiency while achieving stronger attacks. Label flipping is nearly cost-free but ineffective, while the smaller drops on CNN/MLP versus LR indicate that more complex models are less vulnerable.

\begin{table}[!t]
\centering
\caption{Unlearnable examples with different BO algorithms and non-bilevel algorithms. The unlearnable examples trained on CNN and tested on ResNet18 and CNN.}
\label{table:unlearnable-example}
\resizebox{\columnwidth}{!}{%
\begin{tabular}{lcccc}
\toprule
 & ResNet18 & CNN & Time & Bilevel?\\
\midrule
None (clean) & 92.36 & 86.89 & - & - \\
AID-CG       & 78.23 & 57.32 & 11.5h & \Checkmark \\
BOME         & 48.09 & 26.34 & 2.2h & \Checkmark \\
stocBiO      & 84.92 & 75.13 & 6.7h
& \Checkmark \\
AID-KFAC (ours)     & \textbf{\underline{42.33}} & \textbf{\underline{22.96}} & 3.8h & \Checkmark \\
\midrule
EM           & \textbf{32.71} & \textbf{21.83} & $<$0.5h 
& \XSolidBrush \\
\bottomrule
\end{tabular}
}
\end{table}

\paragraph{Unlearnable Examples.} We adopt the GUE framework \citep{liu2024game}, where the attacker is a U-Net trained with a CNN backbone to generate imperceptible perturbations that render training data unexploitable. We evaluate on CIFAR-10 with both CNN and ResNet-18 defenders, comparing KFAC-enhanced bilevel solvers against BOME, stocBiO, AID-CG, and error minimization \citep[EM]{huang2021unlearnable}. Results are shown in \Cref{table:unlearnable-example}. While bilevel approaches can reduce classifier accuracy, they remain computationally expensive. AID-KFAC improves both efficiency and attack strength relative to AID-CG, but EM is still faster and more effective overall. This underscores the strengths and limitations of BO: KFAC narrows the efficiency gap, yet in some large-scale adversarial settings, direct perturbation optimization may remain preferable.

\section{Conclusion}
We introduced KFAC for BO, replacing expensive IHVPs with IKVPs. Across diverse tasks (data hypercleaning, meta-learning for class imbalance, large-scale BERT fine-tuning, continued pretraining, and AI safety problems), our method consistently accelerates convergence and improves performance. Compared to curvature-free methods such as SAMA, our KFAC-based approach preserves richer second-order information with modest overhead while avoiding the prohibitive cost of iterative solvers such as CG and Neumann. These results establish KFAC as a practical and general middle ground, enabling curvature-aware BO at previously infeasible scales. Looking ahead, important directions include developing a theoretical quantification of the extent to which KFAC can benefit BO, refining our approximation by solving the inner problem more efficiently (\eg, \citealt{DongYYZ25}), designing systematic strategies for damping, and exploring alternative structured approximations across broader application domains.

\section*{Acknowledgement}
We gratefully acknowledge funding support from NSERC, the Canada CIFAR AI Chairs program and the Ontario Early Researcher program. Resources used in preparing this research were provided, in part, by the Province of Ontario, the Government of Canada through CIFAR, and companies sponsoring the Vector Institute.
We thank Yihan Wang for some early discussions.

\bibliography{BO-KFAC}
\bibliographystyle{reference}

\clearpage
\section*{Checklist}


\begin{enumerate}

  \item For all models and algorithms presented, check if you include:
  \begin{enumerate}
    \item A clear description of the mathematical setting, assumptions, algorithm, and/or model. [Yes]
    \item An analysis of the properties and complexity (time, space, sample size) of any algorithm. [Yes]
    \item (Optional) Anonymized source code, with specification of all dependencies, including external libraries. [Yes]
  \end{enumerate}

  \item For any theoretical claim, check if you include:
  \begin{enumerate}
    \item Statements of the full set of assumptions of all theoretical results. [Not Applicable] This work is mainly empirical and we make no theoretical claim.
    \item Complete proofs of all theoretical results. [Not Applicable]
    \item Clear explanations of any assumptions. [Not Applicable]     
  \end{enumerate}

  \item For all figures and tables that present empirical results, check if you include:
  \begin{enumerate}
    \item The code, data, and instructions needed to reproduce the main experimental results (either in the supplemental material or as a URL). [Yes]
    \item All the training details (e.g., data splits, hyperparameters, how they were chosen). [Yes]
    \item A clear definition of the specific measure or statistics and error bars (e.g., with respect to the random seed after running experiments multiple times). [Yes]
    \item A description of the computing infrastructure used. (e.g., type of GPUs, internal cluster, or cloud provider). [Yes]
  \end{enumerate}

  \item If you are using existing assets (e.g., code, data, models) or curating/releasing new assets, check if you include:
  \begin{enumerate}
    \item Citations of the creator If your work uses existing assets. [Not Applicable]
    \item The license information of the assets, if applicable. [Not Applicable]
    \item New assets either in the supplemental material or as a URL, if applicable. [Not Applicable]
    \item Information about consent from data providers/curators. [Not Applicable]
    \item Discussion of sensible content if applicable, e.g., personally identifiable information or offensive content. [Not Applicable]
  \end{enumerate}

  \item If you used crowdsourcing or conducted research with human subjects, check if you include:
  \begin{enumerate}
    \item The full text of instructions given to participants and screenshots. [Not Applicable]
    \item Descriptions of potential participant risks, with links to Institutional Review Board (IRB) approvals if applicable. [Not Applicable]
    \item The estimated hourly wage paid to participants and the total amount spent on participant compensation. [Not Applicable]
  \end{enumerate}

\end{enumerate}

\clearpage
\appendix
\crefalias{section}{appendix}
\crefalias{subsection}{appendix}
\onecolumn
\section{Probabilistic Interpretation of Losses and Monte-Carlo KFAC}
\label{app:hessian_sampling}

This section explains how the random vector distribution $p(\vs_n)$ used in the main text arises from the probabilistic interpretation of the loss and how it is instantiated in Monte-Carlo KFAC in practice.

\subsection{How to implement Monte-Carlo KFAC}
\label{app:distribution of p(s)}

Many supervised learning objectives can be written as negative log-likelihoods. Given an input-target pair $(\vx, \vy)$, let the network output be $\vf_\vtheta (\vx) \in \mathbb{R}^C$ that parametrizes a predictive distribution $q (\vy \mid \vf_\vtheta (\vx))$. This viewpoint covers two standard losses used in practice. 

In regression, the loss is
\begin{equation*}
    \ell(\vf, \vy) = \frac{1}{2} \| \vy - \vf \|^2,
\end{equation*}
which corresponds, up to an additive constant, to the negative log-likelihood of the Gaussian model
\begin{equation*}
    q(\vy \mid \vf) = \mathcal{N}(\vy; \vf, \mI).
\end{equation*}

In multiclass classification, if $\vf \in \mathbb{R}^C$ are logits and $\vp = \operatorname{softmax}(\vf)$, then the cross-entropy loss is
\begin{equation*}
    \ell (\vf, \vy) = - \log \vp_\vy
\end{equation*}
is the negative log-likelihood of the categorical model
\begin{equation*}
    q (\vy \mid \vf) = \operatorname{Cat}(\vp).
\end{equation*}

Given the above probabilistic view of loss functions, we now explain how Monte-Carlo KFAC is implemented. For a training example $(\vx_n, y_n)$, let
\[
\vf_n = \vf_\vtheta(\vx_n), \qquad c_n = \ell(\vf_n,\vy_n).
\]
In \Cref{sec:kfac}, we introduced a distribution \(p(\vs_n)\) such that
\[
\nabla_{\vf_n}^2 c_n
=
\mathbb{E}_{\vs_n\sim p(\vs_n)}[\vs_n\vs_n^\top].
\]
This identity says that the Hessian \wrt $\vf_n$, $\nabla_{\vf_n}^2 c_n$, can be represented as the second moment of a suitable random vector \(\vs_n\).
In MC-KFAC, this random vector is constructed by sampling a pseudo-target
\[
\tilde \vy_n \sim q(\cdot \mid \vf_n),
\]
from the model's predictive distribution, and then differentiating the corresponding pseudo-loss
\[
\tilde c_n = -\log q(\tilde \vy_n\mid \vf_n).
\]
The random vector is therefore
\[
\vs_n := \nabla_{\vf_n} \tilde c_n.
\]
Equivalently, \(p(\vs_n)\) is the distribution induced by first drawing \(\tilde \vy_n \sim q(\cdot\mid \vf_n)\), and then mapping it to the gradient
\[
\vs_n = \nabla_{\vf_n}(-\log q(\tilde \vy_n\mid \vf_n)).
\]
With this definition,
\[
\mathbb{E}_{\tilde y_n\sim q(\cdot\mid \vf_n)}[\vs_n\vs_n^\top]
=
\mathbb{E}_{\vs_n\sim p(\vs_n)}[\vs_n\vs_n^\top],
\]
and this expectation equals the output-space GGN term used by KFAC. In practice, we draw $M$ \iid samples $\vs_{n,i} \sim p(\vs_n)$:
\begin{equation*}
    \nabla_{\vf_n}^2 c_n \approx \frac{1}{M} \sum_{i=1}^M \vs_{n,i} \vs_{n,i}^\top.
\end{equation*}
In essence, we applied the following identity: 
\begin{align}
    \mathbb{E}_{\tilde y_n \sim q(\cdot | \vf_n) } [ \nabla_{\vf_n} \log q(\tilde y_n | \vf_n) \cdot \big(\nabla_{\vf_n} \log q(\tilde y_n | \vf_n)\big)^\top ] &= -\mathbb{E}_{\tilde y_n \sim q(\cdot | \vf_n) } [ \nabla^2_{\vf_n} \log q(\tilde y_n | \vf_n) ] 
    \\
    &= - \nabla^2_{\vf_n} \log q( y_n | \vf_n) =: \nabla_{\vf_n}^2 c_n,
\end{align}
where the first equality is a familiar consequence of integration by parts while the second equality holds if the Hessian $\nabla^2_{\vf_n} \log q(y_n | \vf_n)$ does not depend on $y_n$, \eg, for a simple exponential family where 
\begin{align*}
    \log q(y_n | \vf_n) = h(y_n) + T(y_n) \cdot \vf_n - A(\vf_n).
\end{align*}

We now make the distribution \(p(\vs_n)\) explicit for two commonly used loss functions.
\paragraph{Cross-entropy loss.}
Let $\vf_n \in \mathbb{R}^C$ be the logits for example $n$, and let \(\vp_n=\softmax(\vf_n)\), For a lable $y_n \in \{1, \cdots, C\}$, the softmax cross-entropy loss is
\begin{equation*}
    \ell(\vf_n, y_n) = - \log p_{y_n} = -\vf_{n, y_n} + \log \sum_{c=1}^C e^{\vf_{n, c}}. 
\end{equation*}
Its Hessian \wrt $\vf_n$ is
\begin{equation*}
    \nabla_{\vf_n}^2 \ell(\vf_n, y_n) = \operatorname{diag}(\vp_n) - \vp_n \vp_n^\top,
\end{equation*}
which is the standard output-space GGN term for softmax classification \citep{kunstner2019limitations}. To realize this Hessian as a second memoent, we sample a pseudo-target
\begin{equation*}
    \Tilde{y}_n \sim \operatorname{Cat} (\vp_n),
\end{equation*}
and define the corresponding pseudo-loss
\begin{equation*}
    \Tilde{\ell}_n = - \log p_{n, \Tilde{y}_n}.
\end{equation*}
If $\ve_j$ denotes the one-hot vector of class $j$, then when $\Tilde{y}_n = j$,
\begin{equation*}
    \vs_n := \nabla_{\vf_n} \Tilde{\ell}_n = \vp_n - \ve_j.
\end{equation*}
Hence $p(\vs_n)$ is the distribution induced by $j \sim \operatorname{Cat}(\vp)$ and the map
\begin{equation*}
    \vs_n = \vp_n - \ve_j.
\end{equation*}
Since
\begin{equation*}
    \mathbb{E}[\ve_j] = \vp_n, \quad \mathbb{E}[\ve_j \ve_j^\top] = \operatorname{diag}(\vp_n),
\end{equation*}
we obtain that
\begin{equation*}
    \mathbb{E}[\vs_n] = 0,
\end{equation*}
and
\begin{equation*}
\mathbb{E}_{\vs_n \sim p(\vs_n)}[\vs_n \vs_n^\top] = \mathbb{E}[(\vp_n - \ve_j) (\vp_n - \ve_j)^\top] = \operatorname{diag}(\vp_n) - \vp_n \vp_n^\top.
\end{equation*}

Therefore, for cross-entropy, the random vector $\vs_n$ is obtained by sampling a pseudo-label from the model predictive distribution and differentiating the corresponding pseudo-loss, yielding an unbiased rank-one estimator of the output Hessian.

\paragraph{Mean squared error.}
For MSE,
\[
\ell(\vf,\vy)=\frac12\|\vy-\vf\|^2,
\qquad
\nabla_{\vf}^2 \ell(\vf,\vy)=\mI.
\]
In this case, \(p(\vs_n)\) can be taken as a standard Gaussian
\[
\vs_n \sim \mathcal{N}(\mathbf{0},\mI),
\]
so that
\[
\mathbb{E}_{\vs_n\sim p(\vs_n)}[\vs_n\vs_n^\top]=\mI.
\]
Thus, for MSE, the random vector \(\vs_n\) is not obtained by sampling a discrete pseudo-label, but by sampling a Gaussian output perturbation whose covariance matches the output curvature.

\subsection{Weighted cross-entropy and Fisher}
\label{app:weighted_fisher}
In data hypercleaning, each training example $(\vx_n, y_n)$ is assigned a learnable nonnegative weight \(\sigma_n \ge 0\), yielding the weighted objective
\[
\mathcal{L}_{\sigma}(\vtheta)
=
\frac{1}{N}\sum_{n=1}^N \sigma_n\,\ell(\vf_\vtheta(\vx_n),y_n).
\]
For a fixed example \(n\), define the weighted per-example loss
\[
\ell_{n,\sigma}
:=
\sigma_n\,\ell(\vf_n,y_n),
\qquad
\vf_n=\vf_\vtheta(\vx_n).
\]
Since \(\sigma_n\) does not depend on \(\vf_n\), its Hessian with respect to the network output is simply
\[
\nabla_{\vf_n}^2 \ell_{n,\sigma}
=
\sigma_n\,\nabla_{\vf_n}^2 \ell(\vf_n,y_n).
\]

For softmax cross-entropy, let $\vp=\softmax(\vf_n)$. Then the Hessian \wrt $\vf_n$ is
\[
\nabla_{\vf_n}^2 \ell(\vf_n,y_n)
=
\operatorname{diag}(\vp_n)-\vp_n\vp_n^\top,
\]
and therefore the weighted output-space curvature is
\[
\nabla_{\vf_n}^2 \ell_{n,\sigma}
=
\sigma_n\bigl(\operatorname{diag}(\vp_n)-\vp_n\vp_n^\top\bigr).
\]

To realize this matrix as a second moment, we sample a pseudo-label
\[
j \sim \mathrm{Cat}(\vp_n),
\]
and define the vector $\vs_n$
\[
\vs_n
:=
\sqrt{\sigma_n}\,(\vp_n-\ve_j),
\]
where \(\ve_j\) is the one-hot vector for class \(j\). Since
\[
\mathbb{E}_{j\sim \mathrm{Cat}(\vp_n)}
\big[(\vp_n-\ve_j)(\vp_n-\ve_j)^\top\big]
=
\operatorname{diag}(\vp)-\vp\vp^\top,
\]
we obtain
\[
\mathbb{E}[\vs_n\vs_n^\top]
=
\sigma_n\bigl(\operatorname{diag}(\vp_n)-\vp_n\vp_n^\top\bigr)
=
\nabla_{\vf_n}^2 \ell_{n,\sigma}.
\]

Thus, in the weighted setting, the correct Monte-Carlo estimator is obtained by multiplying the usual cross-entropy vector by \(\sqrt{\sigma_n}\). Equivalently, the corresponding pseudo-loss can be written as
\[
\tilde{\ell}_{n,\sigma}
=
\sqrt{\sigma_n}\,(-\log p_{n,j}),
\qquad
j\sim \mathrm{Cat}(\vp_n),
\]
whose gradient with respect to the logits is exactly
\[
\nabla_{\vf_n}\tilde{\ell}_{n,\sigma}
=
\sqrt{\sigma_n}\,(\vp_n-\ve_j).
\]
This ensures that the rank-one estimator \(\vs_n\vs_n^\top\) is unbiased for the weighted output Hessian. Scaling the pseudo-loss directly by \(\sigma_n\) instead would produce
\[
\mathbb{E}[\vs_n\vs_n^\top]
=
\sigma_n^2\bigl(\operatorname{diag}(\vp_n)-\vp_n\vp_n^\top\bigr),
\]
which would over-scale the curvature term.

\section{Related Work}

\subsection{More on KFAC}

Originially introduced by \citet{martens2015optimizing} for MLP, KFAC has been extended to a range of modern architectures, including convolutional networks \citep{grosse2016kroneckerfactored}, recurrent networks \citep{martens2018kroneckerfactored}, and transformers \citep{eschenhagen2023kroneckerfactored}. There are also refinements of the basic approximation, such as eigenvalue-corrected KFAC \citep[EKFAC]{george2018fast}. KFAC has been successfully applied in the context of optimization \citep{osawa2019large}, model sparsification \citep{wang2019eigendamage}, Laplace approximations \citep{daxberger2021laplace} and influence functions \citep{grosse2023studying}.
We are unaware of works applying this curvature approximation in the context of bilevel optimization. 
\subsection{Meta Learning and KFAC}

Several prior works have incorporated curvature information into meta-learning through KFAC or related approximations, but in fundamentally different ways from ours. We will discuss the difference between our work and theirs.

MAML \citep{finn2017model} is a model-agnostic algorithm to train a model with the greatest sensitivity to the loss function of the new task. Suppose there are tasks obeying $T \sim p(T)$ distribution. We can also formulate MAML as a bilevel optimization problem, if we choose the initialization parameter $\vtheta \in \mathbb{R}^d$. 
\begin{equation*}
    \min_\vtheta \mathbb{E}_{T \sim p(T)} [\gL^{\text{val}}(\phi_T^*(\vtheta); T)], \quad \st \quad \phi^*_T(\vtheta) = \argmin_\phi \gL^{\text{tr}} (\phi; T, \vtheta),
\end{equation*}
where $\phi$ is the task-adapted model parameter. That is, we want to find the task-specific parameter $\phi_T^*(\vtheta)$ which will minimize the expected post-adaptation loss across tasks. In practice, MAML does not solve the inner problem to full convergence, but uses one step of SGD from $\vtheta$ to approximate it, and the objective becomes:
\begin{equation*}
    \min_\vtheta \mathbb{E}_{T \sim p(T)} [\gL^{\text{val}} (\vtheta - \alpha \nabla_\vtheta \gL^{\text{tr}} (\vtheta; T), T)],
\end{equation*}

for the learning rate $\alpha$. There are several works that have adapted KFAC to improve the algorithms for MAML. \citet{Zhang2023} replace the inner optimizer with natural gradient descent preconditioned by KFAC, focusing on improving convergence speed rather than on BO. \citet{grant2018recasting} reinterpret MAML as hierarchical Bayes and employ KFAC only to approximate Hessian determinants within Laplace correction to the marginal likelihood, rather than to compute hypergradients. \citet{Kim2023} propose BM-KFP, which applies Kronecker factorization to the parameter of a learned meta-optimizer, a structural efficiency technique unrelated to Hessian approximation. Meta-Curvature \citep{park2019meta} learns gradient transformation matrices for fast adaptation, and to ensure tractability, parameterizes them with low-rank and Kronecker-structured factors, which are conceptually related to KFAC’s block-Kronecker approximation of curvature. Unlike KFAC, however, these operators are not derived from the GGN but are meta-learned directly from data. In contrast, we use KFAC as an efficient means for inverse Hessian-vector products.

\section{Connection between GU and IFT}
\label{app:connection_GU_IFT}
The idea of MAML can be seen as a one-step gradient unrolling (GU), and we can show that GU is equivalent to IFT for the one-step case or the infinite step case. We consider the BO formulation:
\begin{equation*} \min_{\vlambda \in \mathbb{R}^m} \; \Phi(\vlambda) := \mathcal{J}_{\text{out}}\bigl(\vlambda, \vtheta^\star(\vlambda)\bigr), \quad \st \quad \vtheta^\star(\vlambda) = \arg\min_{\vtheta \in \mathbb{R}^d} \mathcal{J}_{\text{in}}(\vlambda,\vtheta). 
\end{equation*}
Here, we demonstrate how to solve it using GU and how this is connected to the hypergradient. Let the inner problem be optimized by $T$ steps of gradient descent with step size $\eta>0$,
\begin{equation}
\label{eq:grad-unroll}
\vtheta^{t+1} \;=\; \vtheta^{t} - \eta\,\nabla_{2}\,\mathcal{J}_{\text{in}}(\vlambda,\vtheta^{t}), 
\qquad t=0,\dots,T-1,
\end{equation}
with initialization $\vtheta^{0}$ (typically independent of $\vlambda$). Define the unrolled outer objective
$\Phi_T(\vlambda):=\mathcal{J}_{\text{out}}(\vlambda,\theta^T(\vlambda))$.
By the chain rule,
\begin{equation}
\label{eq:chain_rule}
\nabla_{\vlambda}\Phi_T(\vlambda)
\;=\;
\nabla_{1}\,\mathcal{J}_{\text{out}}(\vlambda,\vtheta^T)
\;+\;
\bigl(\tfrac{\mathrm{d}\vtheta^T}{\mathrm{d}\vlambda}\bigr)^\top
\,\nabla_{2}\,\mathcal{J}_{\text{out}}(\vlambda,\vtheta^T).
\end{equation}
Let $\mS^t := \tfrac{\mathrm{d}\vtheta^t}{\mathrm{d}\vlambda}\in\mathbb{R}^{d\times m}$ and $\mH_t \coloneqq \nabla_{22}^2 \mathcal{J}_{\text{in}}(\vlambda, \vtheta^t)$. If $\vtheta^0$ does not depend on $\vlambda$ (\ie,  $\mS^0 = \frac{\mathrm{d}\vtheta^0}{\mathrm{d} \vlambda} = 0$), differentiating the update in \cref{eq:grad-unroll} and expanding the recursion, we can get the explicit formula for $\mS^{t+1}$:
\begin{align*}
\mS^{t+1}
 &=
\mS^{t} - \eta\Bigl[\nabla_{21}^2\mathcal{J}_{\text{in}}(\vlambda,\vtheta^{t}) 
\;+\;
\nabla_{22}^2\mathcal{J}_{\text{in}}(\vlambda,\vtheta^{t})\,\mS^{t}\Bigr] \\
&= -\eta \nabla_{21}^2\mathcal{J}_{\text{in}}(\vlambda, \vtheta^t) + (\mI - \eta \mH_t)\mS^t \\
&= -\eta \nabla_{21}^2\mathcal{J}_{\text{in}}(\vlambda, \vtheta^t)  + (\mI - \eta \mH_t ) \big((\mI - \eta\mH_{t-1})\mS^{t-1} - \eta \nabla_{21}^2\mathcal{J}_{\text{in}}(\vlambda, \vtheta^{t-1})\big) \\
&= -\eta \nabla_{21}^2 \mathcal{J}_{\text{in}}(\vlambda, \vtheta^t) - \eta (\mI - \eta \mH_t) \nabla_{21} \mathcal{J}_{\text{in}}(\vlambda, \vtheta^{t-1}) + \big( \prod_{j = t-1}^t (\mI - \eta \mH_j) \big) \mS^{t-1} \\
&= \dots \\
&= -\eta\,\sum_{s=0}^{t}\Biggl(\;\prod_{j=s+1}^{t}\bigl(\mI-\eta \mH_j\bigr)\Biggr)\,\nabla_{21}^2\mathcal{J}_{\text{in}} (\vlambda, \vtheta^s).
\end{align*}
Plugging back to \eqref{eq:chain_rule}, the $T$-step unrolled hypergradient is
\begin{equation*}
\boxed{
\;\nabla_{\vlambda}\Phi_T(\vlambda)
=
\nabla_{1}\,\mathcal{J}_{\text{out}}(\vlambda,\vtheta^T)
\;-\;
\eta\sum_{t=0}^{T-1}
\Bigl[
\nabla_{21}^2 \mathcal{J}_{\text{in}}(\vlambda, \vtheta^t)^\top\,
\Bigl(\prod_{j=t+1}^{T-1}\bigl(\mI-\eta \mH_j\bigr)\Bigr)\,
\nabla_{2}\,\mathcal{J}_{\text{out}}(\vlambda,\vtheta^T)
\Bigr],
\;}
\end{equation*}
where products over an empty index set are the identity.

Now, recall the expression of IFT based hypergradient: 

\begin{equation*}
\boxed{
\nabla_{\vlambda} \Phi(\vlambda) = \nabla_{1}\,\mathcal{J}_{\text{out}}(\vlambda,\vtheta^\star)
\;-\;
\nabla_{12}^2\mathcal{J}_{\text{in}}(\vlambda,\vtheta^\star)
\bigl(\mH\bigr)^{-1}
\nabla_{2}\,\mathcal{J}_{\text{out}}(\vlambda,\vtheta^\star),
}
\end{equation*}
where $\vtheta^*$ is the optimal $\vtheta$ for the inner problem and $\mH = \nabla_{22}^2 \mathcal{J}_{\text{in}}(\vlambda, \vtheta^*)$, and this value is usually estimated by running a few steps of an optimization algorithm. For $T = 1$, we have $\mS^1 = -\eta \nabla_{21}^2 \mathcal{J}_{\text{in}}(\vlambda, \vtheta^0)$, and thus
\begin{equation*}
\;\nabla_{\lambda}\Phi_1(\vlambda)
=
\nabla_{1}\,\mathcal{J}_{\text{out}}(\vlambda,\vtheta^{1})
\;-\;
\eta\,\bigl[\nabla_{21}^2\mathcal{J}_{\text{in}}(\vlambda,\vtheta^{0})\bigr]^\top
\nabla_{2}\,\mathcal{J}_{\text{out}}(\vlambda,\vtheta^{1}).
\end{equation*}
Comparing to the IFT form, if we (i) approximate $\vtheta^* \approx \vtheta^0 \approx \vtheta^1$ and (ii) approximate the inner curvature by the identity,
$\mH \approx \mI$ (so $\mH^{-1}\nabla_{2}\mathcal{J}_{\text{out}}\approx \nabla_{2}\mathcal{J}_{\text{out}}$),
then with $\eta\approx1$ we recover the same algebraic structure:
\begin{equation*}
\nabla_{\vlambda}\Phi(\vlambda)
\;\approx\;
\nabla_{1}\,\mathcal{J}_{\text{out}}(\vlambda,\vtheta^\star)
\;-\;
\nabla_{12}^2\mathcal{J}_{\text{in}}(\vlambda,\vtheta^\star)\,
\nabla_{2}\,\mathcal{J}_{\text{out}}(\vlambda,\vtheta^\star).
\end{equation*}

\citet{lorraine2020optimizing} shows that as $T \rightarrow \infty$, under the assumption $\vtheta^0 = \vtheta^1 = \dots = \vtheta^*$ and $\mI - \eta \mH_t$ is contractive, GU can again happen to align with IFT. To see this, consider $\mS^T$:
\begin{align*}
    \lim_{T \rightarrow\infty} \mS^{T} &= \lim_{T \rightarrow \infty}\Big[ -\eta \sum_{t=0}^{T-1} \big( \prod_{j=t+1}^{T-1} (\mI - \eta \mH_j)\big) \nabla_{21}^2 \mathcal{J}_{\text{in}}(\vlambda, \vtheta^t) \Big]\\
    &= \lim_{T \rightarrow \infty} \Big[-\eta \sum_{t=0}^{T-1}\Big(\prod_{j=t+1}^{T-1}(\mI - \eta \mH)\Big) \nabla_{21}^2 \mathcal{J}_{\text{in}}(\vlambda, \vtheta^*)\Big] \qquad &&\vtheta^0 = \vtheta^* = \vtheta^t \\
    &= \lim_{T \rightarrow \infty} \Big[-\eta\sum_{t=0}^{T-1} \Big(\mI - \eta \mH\Big)^{T-1-t} \nabla_{21}^2 \mathcal{J}_{\text{in}} (\vlambda, \vtheta^*)\Big] \\
    &= \lim_{T \rightarrow \infty} \Big[-\eta\sum_{t=0}^{T-1} \Big(\mI - \eta \mH\Big)^{t} \nabla_{21}^2 \mathcal{J}_{\text{in}} (\vlambda, \vtheta^*)\Big] &&\text{re-index} \\
    &= - \mH^{-1} \nabla_{21}^2 \mathcal{J}_{\text{in}}(\vlambda, \vtheta^*). &&\text{Neumann series and contractive}
\end{align*}
Plugging back to \eqref{eq:chain_rule}, we will get the same expression as the hypergradient. Therefore, under the local optimality and contraction assumption, unrolling to infinity is exactly implicit differentiation: the Neumann series induced by unrolling sums to the inverse Hessian, turning GU into IFT. Our KFAC-based hypergradient can be seen as a middle ground between $\mI$ and $\mH$, balancing the hypergradient accuracy and the computational cost.

\section{BERT Data Hypercleaning}
\label{app:bert_more_explanation}

We provide additional experimental details and results for the BERT data hypercleaning experiments described in the main text. The bilevel formulation is the same as in \Cref{app:Data Hypercleaning}, with the difference that the inner model is a pretrained BERT classifier and we vary the number of trainable layers.

We consider text classification tasks from the WRENCH benchmark \citep{zhang2wrench}, namely \textsc{SemEval}, \textsc{Youtube}, and \textsc{Trec}. These datasets are weak-supervision benchmarks with noisy labeling sources and are large enough to show clear performance differences between bilevel optimization methods while remaining computationally manageable for repeated BERT fine-tuning.

We use the pretrained BERT-base model (110M parameters) from \texttt{HuggingFace}. In the inner problem, we optimize with AdamW using a fixed learning rate of \(10^{-5}\) and a cosine learning-rate scheduler. In the outer problem, we use SGD with momentum \(0.9\) and perform a grid search over learning rates \(\{0.01, 0.1, 1, 10, 100\}\). Each outer iteration contains 10 inner updates. For KFAC and KFAC EMP, we use damping \(10^{-3}\); for CG, we use 3 conjugate-gradient iterations; for Neumann, we use either 3 or 10 truncation steps; and for SAMA, we set \(\alpha=0.1\). All experiments use batch size 128 and maximum sequence length 100. We run each experiment for 2000 iterations and report averages over multiple random seeds on a single NVIDIA RTX 4090 GPU. Full numerical results are shown in \Cref{tab:bert_accuracy}.

\begin{table*}[!ht]
\centering
\renewcommand{\arraystretch}{1}
\caption{Comparison of Methods on \textsc{Trec}, \textsc{SemEval}, and \textsc{Youtube} at Different Tuning Levels. Each entry shows Accuracy and Training Time. Each result is averaged over 5 runs. These are the same results as \Cref{fig:Pareto} but in numbers. }
\label{tab:bert_accuracy}
\begin{tabular}{l|cc|cc|cc}
\toprule
\textbf{Method} & \multicolumn{2}{c|}{\textbf{Trec}} & \multicolumn{2}{c|}{\textbf{SemEval}} & \multicolumn{2}{c}{\textbf{Youtube}} \\
\cmidrule(lr){2-3} \cmidrule(lr){4-5} \cmidrule(l){6-7}
 & Acc(\%) & Time (s) & Acc(\%) & Time (s) & Acc(\%) & Time (s) \\
\midrule
\multicolumn{7}{c}{\textbf{1 Layer (6.5\% Params)}} \\
\midrule
Baseline      & 60.12\(_{\pm0.43}\) & 73.3\(_{\pm0.1}\) & 78.94\(_{\pm0.87}\) & 121.2\(_{\pm0.3}\) & 68.53\(_{\pm1.24}\) & 124.6\(_{\pm1.2}\) \\
KFAC          & \textbf{77.62}\(_{\pm1.28}\) & 772.8\(_{\pm3.4}\) & \textbf{83.83}\(_{\pm0.47}\) & 840.5\(_{\pm3.8}\) & \textbf{81.33}\(_{\pm1.55}\) & 955.7\(_{\pm4.4}\) \\
KFAC EMP      & 74.25\(_{\pm2.57}\) & 791.5\(_{\pm3.5}\) & 81.61\(_{\pm0.75}\) & 838.6\(_{\pm4.5}\) & 80.53\(_{\pm1.32}\) & 962.8\(_{\pm3.2}\) \\
SAMA          & 62.93\(_{\pm0.84}\) & \textbf{602.7}\(_{\pm2.2}\) & 81.61\(_{\pm1.46}\) & \textbf{628.0}\(_{\pm5.7}\) & 77.86\(_{\pm1.15}\) & \textbf{630.1}\(_{\pm5.3}\) \\
CG            & 75.60\(_{\pm1.85}\) & 963.7\(_{\pm3.8}\) & 79.72\(_{\pm0.91}\) & 1020.3\(_{\pm4.5}\) & 77.42\(_{\pm1.58}\) & 1207.9\(_{\pm3.8}\) \\
Neumann ($K{=}3$)  & 53.53\(_{\pm0.75}\) & 1208.8\(_{\pm15.3}\) & 81.22\(_{\pm1.32}\) & 928.5\(_{\pm2.4}\) & 48.89\(_{\pm1.20}\) & 934.7\(_{\pm7.5}\) \\
Neumann ($K{=}10$) & 70.73\(_{\pm3.43}\) & 1681.5\(_{\pm3.6}\) & 82.17\(_{\pm0.36}\) & 1196.8\(_{\pm1.8}\) & 61.47\(_{\pm1.54}\) & 1198.0\(_{\pm3.3}\) \\
\midrule
\multicolumn{7}{c}{\textbf{7 Layers (45.3\% Params)}} \\
\midrule
Baseline      & 63.43\(_{\pm0.68}\) & 135.9\(_{\pm0.1}\) & 79.89\(_{\pm0.42}\) & 152.8\(_{\pm0.1}\) & 69.73\(_{\pm0.19}\) & 162.3\(_{\pm3.6}\) \\
KFAC          & \textbf{73.42}\(_{\pm1.08}\) & 1414.3\(_{\pm2.6}\) & \textbf{85.61}\(_{\pm0.08}\) & 1475.2\(_{\pm2.9}\) & \textbf{79.92}\(_{\pm1.94}\) & 1491.0\(_{\pm6.9}\) \\
KFAC EMP      & 72.17\(_{\pm1.09}\) & 1442.1\(_{\pm3.0}\) & 85.02\(_{\pm1.03}\) & 1475.4\(_{\pm2.0}\) & 79.24\(_{\pm0.82}\) & 1504.3\(_{\pm5.4}\) \\
SAMA          & 63.07\(_{\pm0.52}\) & \textbf{828.3}\(_{\pm4.1}\) & 80.78\(_{\pm0.64}\) & \textbf{878.4}\(_{\pm7.3}\) & 73.43\(_{\pm1.68}\) & \textbf{873.1}\(_{\pm11.9}\) \\
CG            & 68.27\(_{\pm0.50}\) & 2285.6\(_{\pm12.2}\) & 83.67\(_{\pm0.62}\) & 2332.4\(_{\pm7.4}\) & 72.43\(_{\pm1.39}\) & 1967.3\(_{\pm6.2}\) \\
Neumann ($K{=}3$)  & 61.07\(_{\pm0.34}\) & 2177.0\(_{\pm5.5}\) & 78.61\(_{\pm1.13}\) & 2195.4\(_{\pm9.9}\) & 67.62\(_{\pm4.27}\) & 1897.2\(_{\pm7.8}\) \\
Neumann ($K{=}10$) & 64.87\(_{\pm2.86}\) & 4336.1\(_{\pm8.8}\) & 80.33\(_{\pm2.43}\) & 4327.5\(_{\pm10.9}\) & 64.13\(_{\pm4.15}\) & 4375.6\(_{\pm31.0}\) \\
\midrule
\multicolumn{7}{c}{\textbf{12 Layers (100\% Params)}} \\
\midrule
Baseline      & 60.58\(_{\pm0.59}\) & 189.6\(_{\pm0.2}\) & 80.06\(_{\pm1.14}\) & 190.6\(_{\pm0.1}\) & 67.07\(_{\pm0.50}\) & 184.8\(_{\pm0.1}\) \\
KFAC          & 63.67\(_{\pm1.43}\) & 1988.1\(_{\pm2.9}\) & \textbf{82.72}\(_{\pm1.51}\) & 1986.9\(_{\pm5.4}\) & \textbf{76.08}\(_{\pm1.94}\) & 2110.8\(_{\pm5.5}\) \\
KFAC EMP      & \textbf{63.74}\(_{\pm1.17}\) & 2027.6\(_{\pm3.1}\) & 82.56\(_{\pm0.83}\) & 2020.9\(_{\pm2.6}\) & 75.80\(_{\pm1.28}\) & 2117.2\(_{\pm6.4}\) \\
SAMA          & 59.80\(_{\pm0.65}\) & \textbf{1038.7}\(_{\pm6.1}\) & 78.50\(_{\pm0.89}\) & \textbf{1096.2}\(_{\pm3.6}\) & 71.07\(_{\pm0.94}\) & \textbf{1096.2}\(_{\pm3.6}\) \\
CG            & 62.79$_{\pm1.61}$ & 4609.8$_{\pm4.3}$ & 82.32$_{\pm1.23}$ & 4618.3$_{\pm5.2}$ & 73.56$_{\pm1.15}$ & 4607.2$_{\pm5.9}$ \\
Neumann ($K{=}3$)  & 61.42$_{\pm2.01}$ & 3823.2$_{\pm6.2}$ & 81.92$_{\pm0.92}$ & 3828.4$_{\pm5.3}$ & 71.78$_{\pm1.42}$ & 3825.4$_{\pm1.37}$ \\
Neumann ($K{=}10$) & 61.53$_{\pm1.75}$ & 8729.2$_{\pm6.3}$ & $82.53_{\pm1.24}$ & 8727.9$_{\pm5.7}$ & 72.34$_{\pm{1.24}}$ & 8735$_{\pm6.2}$\\
\bottomrule
\end{tabular}
\end{table*}

\section{Data Hypercleaning}
\label{app:Data Hypercleaning}

\begin{figure}[!ht]
    \centering
    \includegraphics[width=1\linewidth]{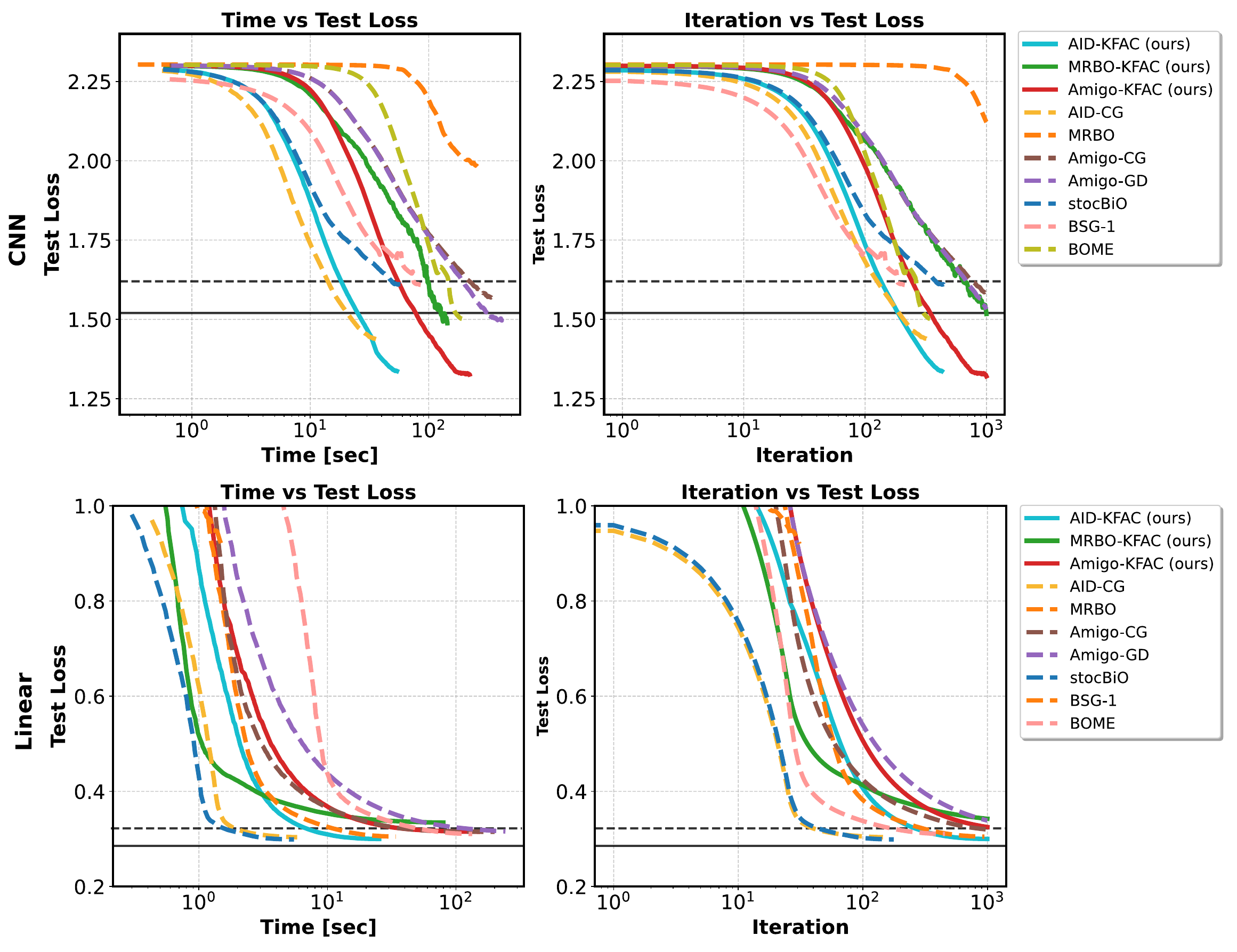}
    \caption{Data hypercleaning results for CNN on CIFAR-10 and linear on MNIST. For the linear model, Solid black line is the test loss reached on the cleaned dataset, while the dashed black line is trained only on validation set. For CNN, solid black line is test loss reached on validation dataset, while dashed line is trained on the validation dataset and noisy dataset.}
    \label{fig:data hypercleaning full}
\end{figure}

 The training dataset, denoted as \(\dataset{tr} = \{(\vx_i, y_i)\}_{i=1}^m\), is corrupted by randomly assigning labels to a proportion of the data points. The objective is to assign weights to the training data points such that a model trained on the weighted training set performs optimally on the validation dataset \(\dataset{val}\). This process can be formulated as:
\begin{equation*}
\min_{\vlambda, \vtheta} \mathcal{L}^{\text{val}}(\vtheta^\star), \quad ~\text{s.t.} ~ \quad \vtheta^\star \!\in\! \argmin_{\vtheta} \left\{ \mathcal{L}^{\text{tr}}(\vtheta, \vlambda) \!+\! \alpha\|\vtheta\|^2 \right\},
\end{equation*}
where \(\mathcal{L}^{\text{val}}(\vtheta^\star)\) is the cross-entropy loss evaluated on the validation set. \(\mathcal{L}^{\text{tr}}(\vtheta, \vlambda)\) is the weighted training loss, defined as:
$\mathcal{L}^{\text{tr}}(\vtheta, \vlambda) = \sum_{n=1}^N \sigma(\lambda_n) \ell(\vx_n, y_n, \vtheta),
$where \(\sigma(\lambda_n) = \text{Clip}(\lambda_n, [0, 1])\) ensures that the weights \(\lambda_n\) remain in the range \([0, 1]\).
The training loss \(\ell(\vx_n, y_n, \vtheta)\) is also computed using cross-entropy. The term \(\alpha \|\vtheta\|^2\) adds an \(L_2\) regularization to prevent overfitting.
The goal is to optimize both the data weights \(\vlambda\) and the model parameters \(\vtheta\) to achieve the best validation performance.

The overall structure of our code was adapted from \url{https://github.com/Cranial-XIX/BOME}. However, in BOME's implementation, their model was implemented as a tensor using Pytorch. In our implementation, we implemented LR, MLP and CNN using \texttt{torch.nn.Module}, which is more practical. However, this will slow down first-order methods like BSG-1 and BOME since we need to convert each layer of the model to a vector.

In the experiment, we split the dataset (both MNIST and CIFAR-10) into 4 parts. The training data size is 50000, and the validation data 1 size is 5000, the validation data 2 size is 5000, and the test data size is 10000. We train the models on training data, update the weight $v$ according to validation data 1, and tune the other hyperparameters on validation data 2. We train on MNIST (linear) and CIFAR-10 (three-layer CNN, ResNet-18) with 50\% label corruption, and we report the results on the test data. We fix the number of inner steps to 10. For the step sizes, we conducted a grid search over inner learning rate $\eta_\vtheta \in \{0.1, 0.01, 0.001\}$ and outer learning rate $\eta_\mathbf{w} \in \{0.1, 1, 10, 100, 500, 1000 \}$, and whether to apply momentum by searching the momentum value over \{0, 0.9\}. For BOME, BVSFM and BSG-1, we also search their hyperparameter over \{0.001, 0.01, 0.1, 1\}. For KFAC, we also grid search the damping value over \{0.001, 0.01, 0.1, 1, 10, 100\}. We report the results for MNIST and CIFAR in \Cref{fig:data hypercleaning full}. We can see that our method not only results in a faster convergence rate but also helps to converge to a better optimum point.

\section{More Applications}
\label{app:more application details}

In this section, we introduce the detailed problem formulations and experimental details for the four applications we consider. For the class imbalance and the continued pretraining problem, our experiment code is based on the \texttt{Betty}  library \citep{choe2022betty} , which provides a template to add more hypergradient estimation methods.

\subsection{Class Imbalance}

Following \citet{shu2019meta}, we aim to train a Meta-Weight-Net (MWN) that assigns weights to different training samples to achieve good performance on the validation set. We can formulate the problem as:
\begin{equation*}
    \min_{\vlambda} \mathcal{L}^{\text{meta}}(\vtheta^*), ~\st~ \vtheta^* \in \argmin_{\vtheta} \sum_{n} w(\ell_n (\vx_n, y_n, \vtheta), \vlambda) \cdot \ell_n (\vx_n, y_n, \vtheta), 
\end{equation*}
where $w(\cdot, \vlambda)$ is MWN parametrized by $\vlambda$,  taking the training loss of the $n$-th data point, and returning a weight. The $\mathcal{L}^{\text{meta}}$ is the loss on a small meta-set. We construct the long-tailed CIFAR dataset by reducing the number of training samples per class using an exponential function $n = n_i \mu^i$, where $i$ is the class index, $n_i$ is the original number of training samples. We train ResNet-32 with softmax cross-entropy loss and SGD, with momentum 0.9, a weight decay of $5 \times 10^{-4}$ and a learning rate 0.1.  The outer learning rate is set to $10^{-5}$ with the Adam optimizer. For SAMA and KFAC, we train the inner problem for 10 steps per outer iteration and set SAMA $\alpha = 0.1$ and damping value for KFAC to be $10^{-3}$. We train the model for 100 epochs, and we randomly select 10 images per class on the validation set as the meta dataset. For the $T_1 - T_2$ method, we use the code in \url{https://github.com/xjtushujun/Meta-weight-net_class-imbalance}.

\subsection{Continued Pretraining}

We further evaluate our method on the continued pretraining task for large language models, following \citet{dery2021should} and \citet{choe2023making}. Continued pretraining aims to adapt a pretrained language model to domain-specific data while avoiding negative transfer from low-quality samples. \citet{choe2023making} formulate the continued pretraining task as BO:

\begin{equation*}
\min_{\vlambda}
\mathcal{L}^{\text{ft}}(\mathcal{D}_{\text{ft}}, \vtheta^*),
\quad \text{s.t.} \quad
\vtheta^* =
\argmin_{\vtheta}
\Big[
\mathcal{L}^{\text{ft}}(\mathcal{D}_{\text{ft}}, \vtheta)
+
\frac{1}{|\mathcal{D}_{\text{pt}}|}
\sum_{\vx \in \mathcal{D}_{\text{pt}}}
w(\vx, \vlambda) \mathcal{L}^{\text{pt}}(\vx; \vtheta)
\Big],
\end{equation*}
where $\mathcal{L}^{\text{ft}}$ and $\mathcal{L}^{\text{pt}}$ are finetuning and pretraining loss functions (respectively), $\mathcal{D}_{\text{ft}}$ and $\mathcal{D}_{\text{pt}}$ are finetuning and pretraining datasets (respectively), and $w(\cdot, \vlambda)$ is the MWN similar to class imbalance, which dynamically reweighs samples from the pretraining corpus $\mathcal{D}_{\text{pt}}$, mitigating the negative effect of noisy or low-quality data.

We use RoBERTa-base as the backbone for both the downstream and auxiliary pretraining tasks; each task is optimized with Adam at an initial learning rate of $2\times10^{-5}$ and a linear decay schedule with 0.6 warmup proportion. No weight decay is applied. The meta-network, implemented as a two-layer MLP, is optimized with Adam at a learning rate of $1\times10^{-5}$ without a scheduler. For SAMA and our KFAC-integrated variant, we use 10 inner steps per outer iteration and set SAMA $\alpha = 0.3$ and KFAC damping value to be $10^{-3}$. All experiments were conducted on the ChemProt, HyperPartisan, ACL-ARC, and SciERC datasets, with a batch size of 16 and a maximum sequence length of 256.

\subsection{Data Poisoning}

Data poisoning attacks aim to inject a small fraction of poisoned samples into the training dataset, thereby degrading the performance of the model trained on the corrupted data. As described by \citet{lu2022indiscriminate}, this problem can be expressed as a BO problem:
\[
\max_{\dataset{p}} \mathcal{L}^{\text{val}}(\dataset{val}, \vtheta^*), \quad 
\text{s.t.} \quad \vtheta^* \in \argmin_{\vtheta} \mathcal{L}^{\text{tr}}(\dataset{tr} \cup \dataset{p}, \vtheta),
\]
Where $\mathcal{L}^{\text{tr}}$ and $\mathcal{L}^{\text{val}}$ represent the training and validation losses, which can be implemented as the cross-entropy loss, \(\dataset{p}\) denotes the poisoned data injected into the training set. The goal is to optimize the poisoned data \(\dataset{p}\) such that a model trained on the combined dataset \(\dataset{tr} \cup \dataset{p}\) performs poorly on the validation dataset \(\dataset{val}\). The implementation of TGDA was adapted from the repo: \url{https://github.com/watml/TGDA-Attack}.

During training time, we use generator models (attackers) to generate poisoned samples from random noise. The attacker model for the MNIST dataset is a three-layer neural network, with three fully connected layers and leaky ReLU activations; for the CIFAR-10 dataset, we use an autoencoder with three convolutional layers as encoder and three conv transpose layers as decoder. For the classifier on the MNIST dataset, we used logistic regression (LR) and Multi-layer Perceptron (MLP), while on the CIFAR-10 dataset, we used a convolutional neural network (CNN) with two convolutional layers, maxpooling and one fully connected layer.

In the data poisoning training setup, we used a batch size of 1000 for all methods on the MNIST dataset. Even for deterministic methods such as BOME AID-CG, batch processing was employed due to the computational expense of alternative approaches. For AID-KFAC, the damping value was set to $10^{-3}$. The models were trained for 200 epochs, with the attacker's learning rate set to 0.1 and the classifier's learning rate set to 0.01. Following \citet{lu2022indiscriminate}, for each update of the attacker, the classifier was updated 20 times. On the CIFAR-10 dataset, we used a batch size of 256 for all methods and trained for 200 epochs. The learning rates were set to 0.1 for the upper level (attacker) and 0.01 for the classifier. Both the attacker and classifier optimizers utilized SGD. For stocBiO, we set the Neumann series parameter $K = 3$. For BOME, the upper-level learning rate was 0.1, while the lower-level learning rate was 0.01. As outlined by \citet{lu2022indiscriminate}, pretraining the attacker and classifier was necessary to ensure convergence. The attacker was pretrained using SGD for 100 epochs with a learning rate of 0.1, utilizing MSE reconstruction loss. Similarly, the classifier was pretrained using SGD with a learning rate of 0.1 for 100 epochs.


\subsection{Unlearnable Example}
\label{unlearnable example detail}

We follow the problem formulation of GUE  \citep{liu2024game}. On a  dataset $\dataset{tr} = \{(\vx_i, y_i)\}_{i=1}^m$, an attacker (leader) employs a generator model to produce perturbations, and a classifier (follower) mimics the victim, with only access to the poisoned data, namely clean data plus the imperceptible perturbations. The classifier minimizes the cross-entropy loss on the poisoned data, but the resulting model exhibits poor performance on clean validation data. Formally, the objective is as follows:
\begin{align*}
    \min_{\vlambda, \vtheta} \mathcal{L}^{\text{val}}(\vtheta), \quad \st \quad \vtheta^* \in \argmin_{\vtheta} \mathcal{L}^{\text{tr}}(\vtheta, \vlambda),
\end{align*}
where $\mathcal{L}^{\text{val}}(\vtheta) = - \frac{1}{m}\sum_{n} \ell_n(\vx_n, y_n, \vtheta)$, $\mathcal{L}^{\text{tr}}(\vtheta, \vlambda) = \frac{1}{m} \sum_{n} \ell_n(\vx_n + g_{\vlambda} (\vx_n), y_n; \vtheta)$, $\ell_n$ is the cross-entropy loss, and $g_{\vlambda}$ represents the generator that produces small perturbations for each training example $\vx_i$. To address gradient explosion issues caused by the unbounded outer objective function, \citet{liu2024game} proposed replacing $\mathcal{L}^{\mathrm{val}}$ by the following surrogate loss during training, for each pair of samples $(\vx, y)$:
\begin{align*}
    \mathcal{L}(\vx, y; \vtheta) \coloneqq \max_{y' \neq y} \ell_n(\vx, y'; \vtheta).
\end{align*}
Minimizing this surrogate loss means minimizing the cross-entropy  with respect to all other labels instead of $y$. \citet{liu2024game} also showed this loss is bounded and has the same monotonicity as regular cross-entropy.

The framework implementation and BOME implementation were adapted from the repo: \url{https://github.com/hong-xian/gue/tree/master/attacks}, while the implementation of AID-CG and AID-KFAC is the same as in data poisoning.

The dataset is CIFAR-10, with 50000 clean training samples and 10000 test samples. For unlearnable examples, we use generator models to generatee a small perturbation for each input image. We used the U-Net as a poison generator during training. We clip the perturbation radius to $\epsilon = 25/255$, since below this threshold we found that most BO algorithms fail to generate effective perturbations that can significantly downgrade the performance of the classifier. For the classifier model, we used a three-layer CNN with ReLU activation and two fully connected layers.

During training, we train 50 epochs for attackers which generate small perturbations. We used a batch size of 512 for training, with the attacker's learning rate set to 1 and the classifier's learning rate set to 0. We used the SGD optimizer for both the attacker and classifer. And we update the attacker 10 times after updating the classifier 10 times. However, we found that for BOME, the batch size 512 fails to generate perturbations that can significantly downgrade the performance of the classifier, so we used batch size 256 for BOME. For the inner approximation of BOME, we used Adam \citep{kingma2014adam} with a learning rate of 0.001 for $T = 10$ iterations. For evaluation, we train the model on a poisoned dataset for 100 epochs using an SGD optimizer with an initial learning rate of 0.01 and test the classification accuracy on the test set.

\section{Ablation Study}
\subsection{Exponential Moving Average (EMA)}
\label{app:EMA}

\begin{figure}[!t]
    \centering
    \begin{subfigure}{0.45\textwidth}
        \centering
        \includegraphics[width=\linewidth]{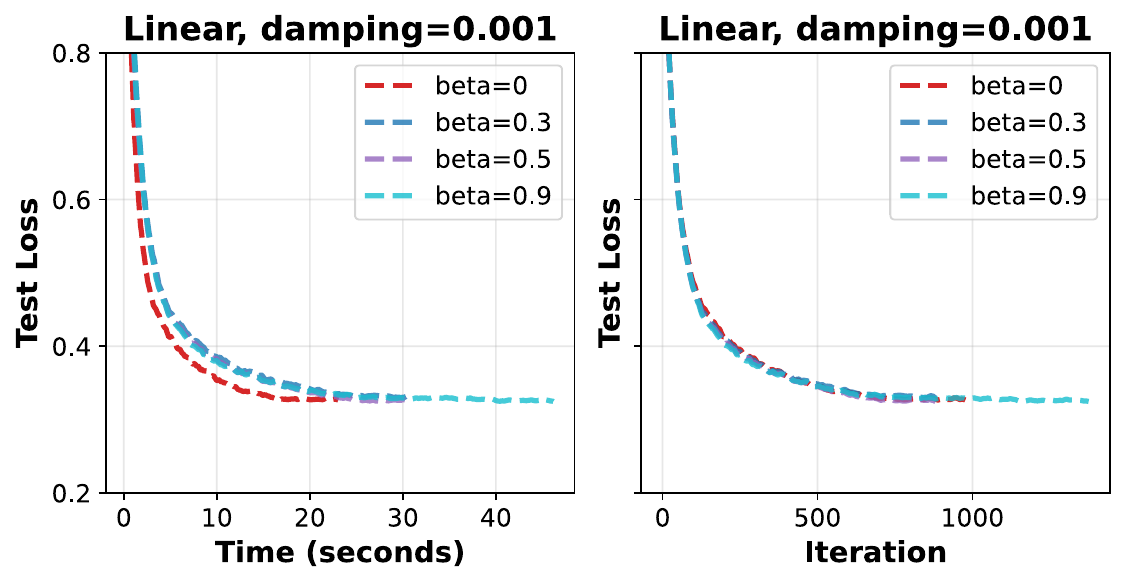}
        \caption{damping = 0.001}
    \end{subfigure}
    \hfill
    \begin{subfigure}{0.45\textwidth}
        \centering
        \includegraphics[width=\linewidth]{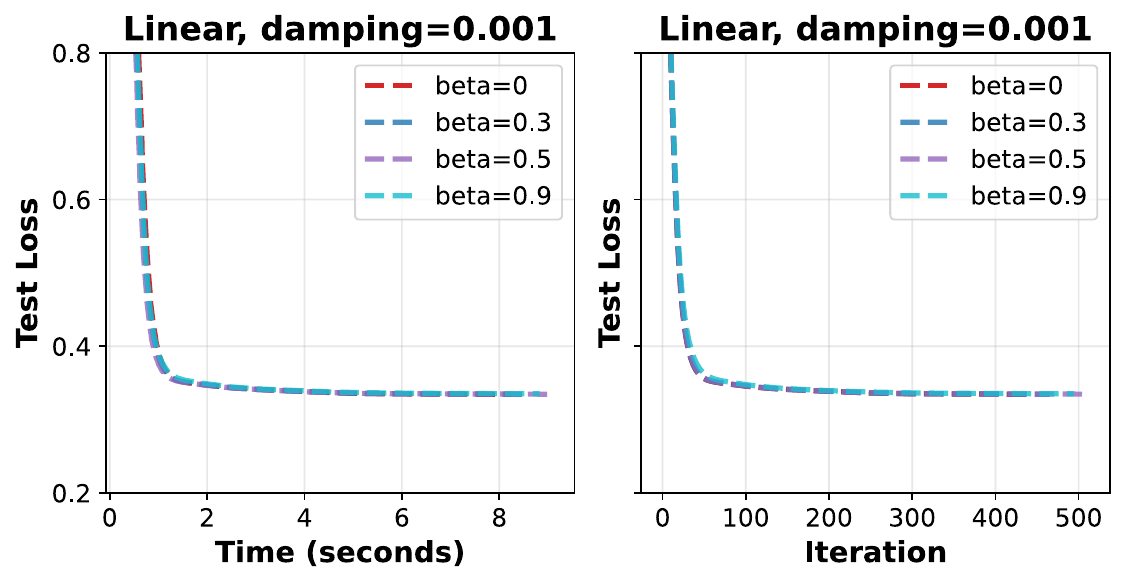}
        \caption{damping = 0.001}
    \end{subfigure}

    \vskip\baselineskip
    \begin{subfigure}{0.45\textwidth}
        \centering
        \includegraphics[width=\linewidth]{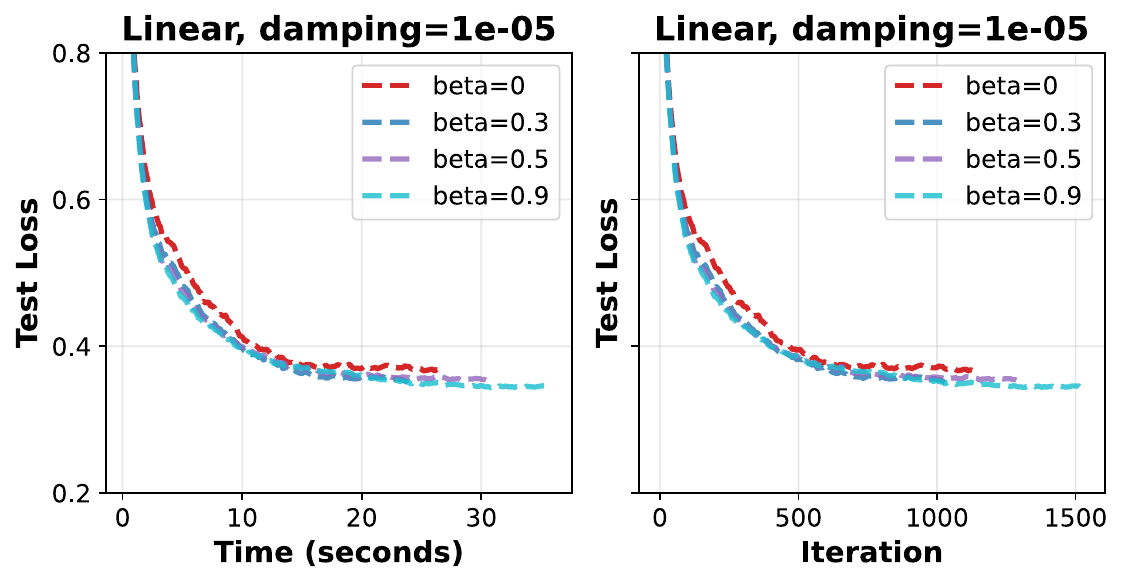}
        \caption{damping = $10^{-5}$}
    \end{subfigure}
    \hfill
    \begin{subfigure}{0.45\textwidth}
        \centering
        \includegraphics[width=\linewidth]{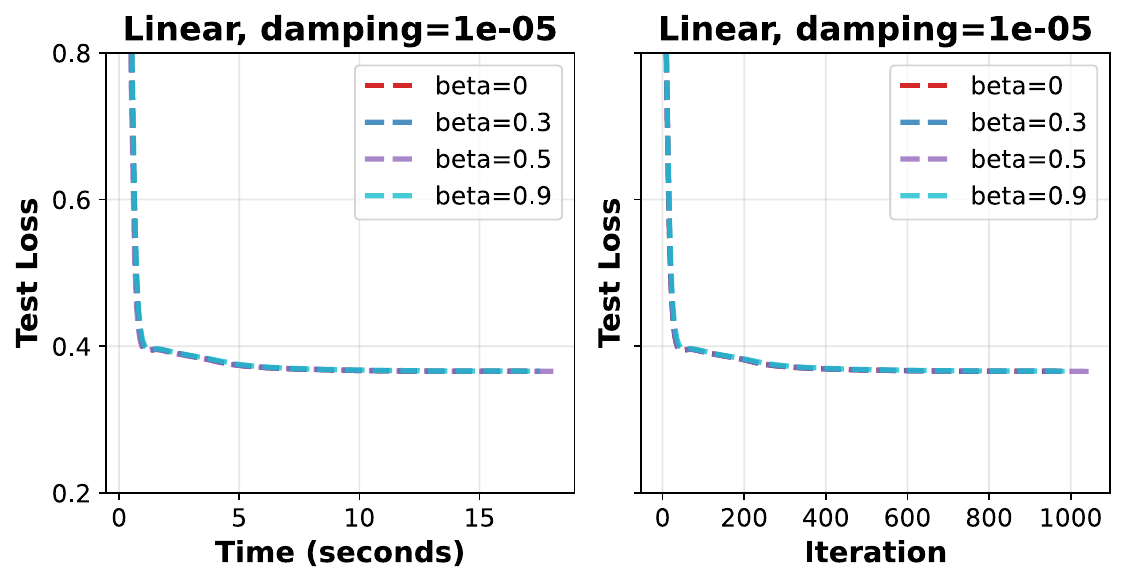}
        \caption{damping = $10^{-5}$}
    \end{subfigure}

    \vskip\baselineskip
    \begin{subfigure}{0.45\textwidth}
        \centering
        \includegraphics[width=\linewidth]{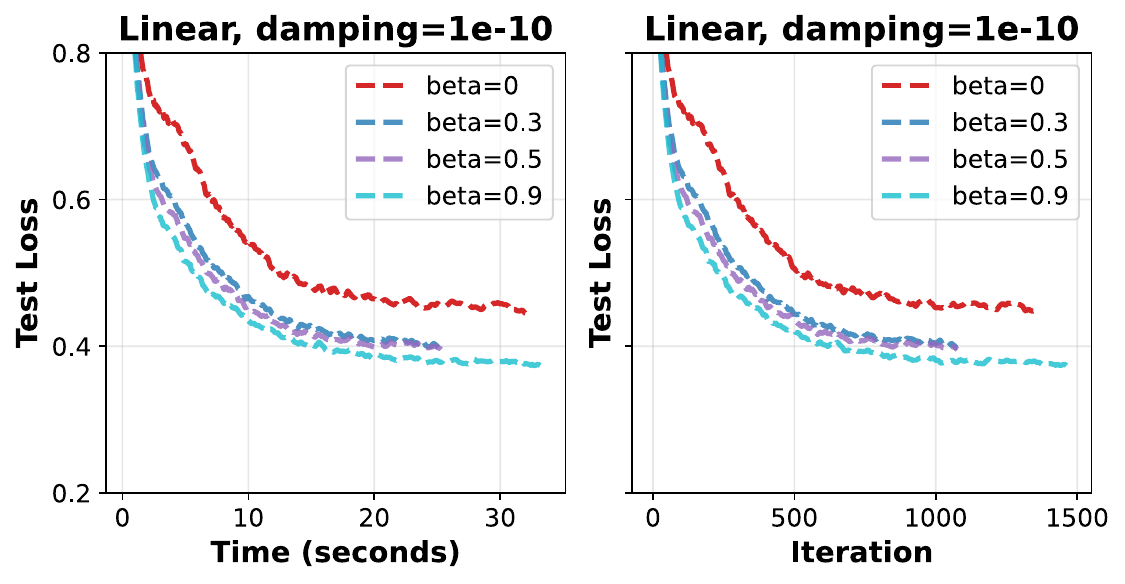}
        \caption{damping = $10^{-10}$}
    \end{subfigure}
    \hfill
    \begin{subfigure}{0.45\textwidth}
        \centering
        \includegraphics[width=\linewidth]{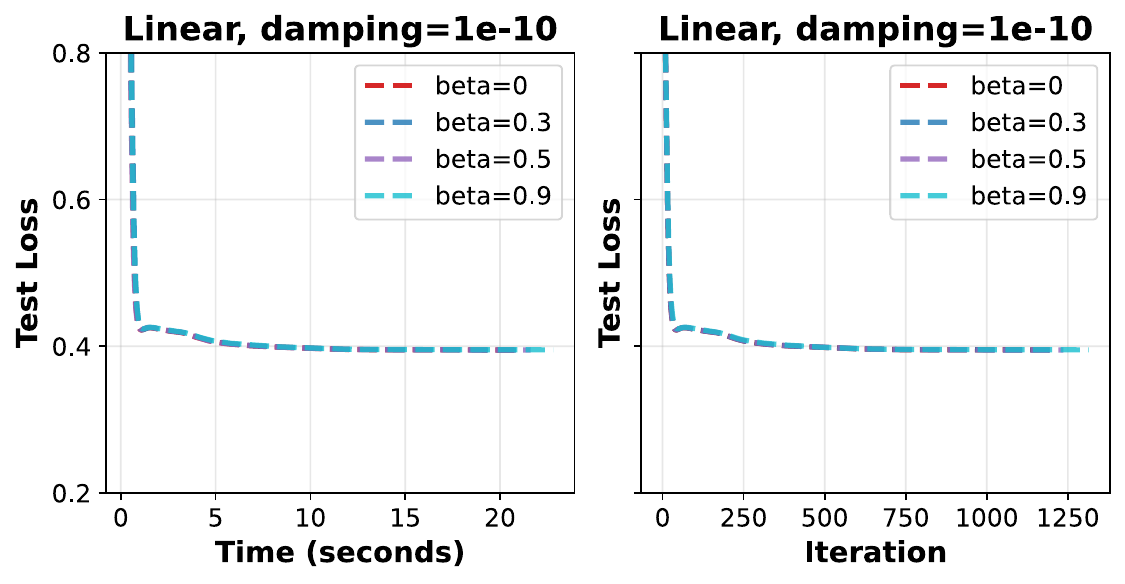}
        \caption{damping = $10^{-10}$}
    \end{subfigure}

\caption{
Effect of EMA coefficient \(\beta\) on convergence for data hypercleaning with logistic regression on MNIST.
Each curve corresponds to a different \(\beta \in \{0, 0.3, 0.5, 0.9\}\).
\textbf{Left column}: batch size 1000. \textbf{Right column}: batch size 50000.
In this setting, EMA is most helpful for small damping and small batch sizes, where curvature estimates are noisier, while its effect is modest for larger batches or stronger damping.}
\label{fig:ema_effect}
\end{figure}

\citet{martens2015optimizing} suggest maintaining running estimates of
$\mA_{\text{KFAC}}$ and $\mB_{\text{KFAC}}$ using an exponential moving average (EMA),
which can smooth stochastic curvature estimates across iterations. After computing
new mini-batch estimates $\mA_{\text{KFAC}}^{\text{new}}$ and
$\mB_{\text{KFAC}}^{\text{new}}$, the running factors are updated as
\begin{equation*}
    \mA_{\text{KFAC}} \leftarrow \beta \mA_{\text{KFAC}} + (1-\beta)\mA_{\text{KFAC}}^{\text{new}},
    \qquad
    \mB_{\text{KFAC}} \leftarrow \beta \mB_{\text{KFAC}} + (1-\beta)\mB_{\text{KFAC}}^{\text{new}},
\end{equation*}
where $\beta\in[0,1)$ controls the amount of temporal smoothing.

To study the role of EMA in BO, we evaluate data hypercleaning with logistic regression on MNIST,
varying the batch size \(B\in\{1000,50000\}\), the EMA coefficient
\(\beta\in\{0,0.3,0.5,0.9\}\), and the damping strength. The results are shown in
\Cref{fig:ema_effect}. In this setting, EMA is most beneficial when both the batch size
and damping are small, where curvature estimates are noisier and temporal averaging can
stabilize the KFAC factors. By contrast, for larger batches or stronger damping, the
performance differences across \(\beta\) are small, suggesting that EMA provides only
limited additional benefit once the curvature estimates are already stable.

Overall, these results support EMA as a useful stabilization mechanism in noisier BO
regimes, rather than as a uniformly necessary component. Since our main experiments do
not operate in this particularly noisy small-batch regime, we use the simpler
non-EMA variant there. We note, however, that these conclusions are drawn from a
small-scale linear model, and EMA may still be valuable in larger or more stochastic
settings.
\clearpage

\subsection{Batch Size}
\label{app:batch size}
\begin{figure}[!h]
    \centering
    \includegraphics[width=1\linewidth]{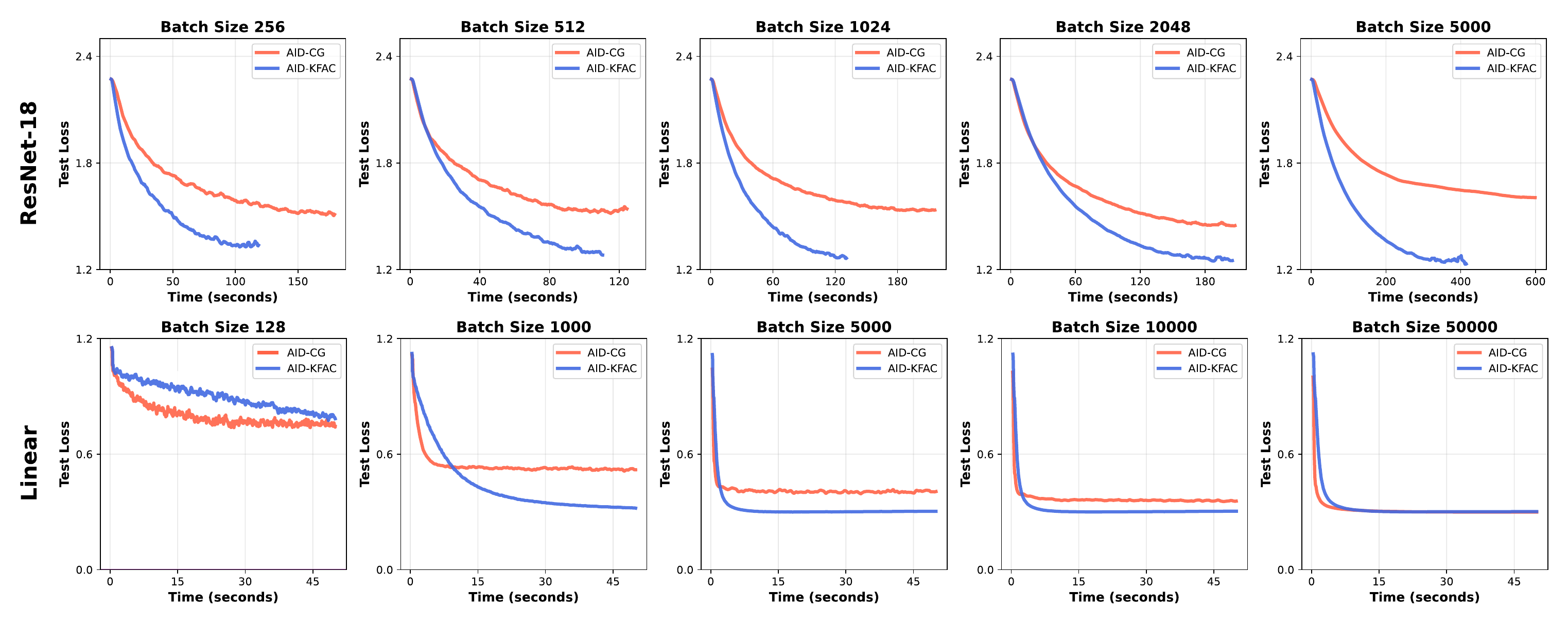}
    \caption{Convergence of AID-CG and AID-KFAC on data hypercleaning for different batch sizes.  
    KFAC achieves faster and more stable convergence, particularly for smaller batches, where curvature estimates are noisier and the inner problem becomes ill-conditioned. Compared with \Cref{fig:batch size study}, this figure shows the evolution of the test loss for different batch sizes.}
    \label{fig:convergence}
\end{figure}

We compare AID-CG and AID-KFAC under varying batch sizes, following the same grid search setup described in \Cref{app:Data Hypercleaning}.  
Specifically, we consider batch size $B= \{256, 512, 1024, 2048, 5000\}$ for ResNet-18 and $B= \{128, 1000, 5000, 10000, 50000\}$ for the linear model and consider the lowest test loss achieved within 1000 optimization iterations. We report the extra results in \Cref{fig:convergence}. The results show that batch size has a noticeable effect on BO performance. In both the ResNet-18 and linear settings, AID-KFAC consistently matches or outperforms AID-CG across all tested batch sizes, with the largest gains typically appearing in the smaller-batch regime. As the batch size increases, the performance gap narrows, especially for the linear model, but AID-KFAC remains competitive throughout. These results suggest that KFAC provides a robust curvature surrogate across a wide range of practical batch-size settings.

\end{document}